\documentclass[conference]{IEEEtran}

\usepackage{times}

\usepackage[font={small}]{caption}
\usepackage{mwe}
\usepackage[numbers]{natbib}
\usepackage{multicol}
\usepackage[bookmarks=true]{hyperref}
\usepackage{color}
\let\labelindent\relax
\usepackage{enumitem}
\usepackage{graphicx}
\usepackage{amsmath}
\usepackage{
tikz,
relsize,
booktabs
}
\usepackage{algorithmic}

\renewcommand\thefigure{\arabic{figure}}
\newcommand{\xxnote}[3]{}
\ifx\hidenotes\undefined
  \renewcommand{\xxnote}[3]{\color{#2}{#1: #3}}
\fi

\hypersetup{
    colorlinks=true,
    linkcolor=blue,
    filecolor=magenta,      
    citecolor=blue,
    urlcolor=teal,
}

\newcommand{\method}{\textsc{T-Dex}}
\newcommand{\website}{\url{tactile-dexterity.github.io}}

\begin{document}

\title{Dexterity from Touch: Self-Supervised Pre-Training of Tactile Representations with Robotic Play}

\author{\authorblockN{Irmak Guzey}
\authorblockA{New York University}
\and
\authorblockN{Ben Evans}
\authorblockA{New York University}
\and
\authorblockN{Soumith Chintala}
\authorblockA{Meta AI}
\and
\authorblockN{Lerrel Pinto}
\authorblockA{New York University}
}

\maketitle

\begin{abstract}
Teaching dexterity to multi-fingered robots has been a longstanding challenge in robotics. Most prominent work in this area focuses on learning controllers or policies that either operate on visual observations or state estimates derived from vision. However, such methods perform poorly on fine-grained manipulation tasks that require reasoning about contact forces or about objects occluded by the hand itself. In this work, we present \method{}, a new approach for tactile-based dexterity, that operates in two phases. In the first phase, we collect 2.5 hours of play data, which is used to train self-supervised tactile encoders. This is necessary to bring high-dimensional tactile readings to a lower-dimensional embedding. In the second phase, given a handful of demonstrations for a dexterous task, we learn non-parametric policies that combine the tactile observations with visual ones. Across five challenging dexterous tasks, we show that our tactile-based dexterity models outperform purely vision and torque-based models by an average of 1.7X.
Finally, we provide a detailed analysis on factors critical to \method{} including the importance of play data, architectures, and representation learning.
\end{abstract}

\IEEEpeerreviewmaketitle

\section{Introduction}

Humans are able to solve novel and complex manipulation tasks with small amounts of real-world experience. Much of this ability can be attributed to our hands, which allow for redundant contacts and multi-finger manipulation. Endowing multi-fingered robotic hands such dexterous capabilities has been a long-standing problem, with approaches ranging from physics-based control~\cite{kumar2016dext} to simulation to real (sim2real) learning~\cite{openai2019learning, Openai2019}. More recently, the prevalence of improved hand-pose estimators has enabled imitation learning approaches to teach dexterity, which in turn improves sample efficiency and reduces the need for precise object and scene modelling~\cite{arunachalam2022dexterous, holodex, DexPilot}.

Even with improved algorithms, teaching dexterous skills is still quite inefficient, requiring hours of demonstration data for imitation or days of training for sim2real~\cite{holodex, openai2019learning}. While algorithmic improvements in control will inevitably lead to improvements in dexterity over time, an often overlooked source of improvement lies in the sensing modality. Current dexterous robots either use high-dimensional visual data or compact states estimated from them. Both suffer significantly either when the task requires reasoning about contact forces, or when the fingers occlude the object being manipulated. In contrast to vision, tactile sensing provides rich contact information while being unaffected by occlusions.

\begin{figure}[t]
    \centering
    \includegraphics[width=\linewidth]{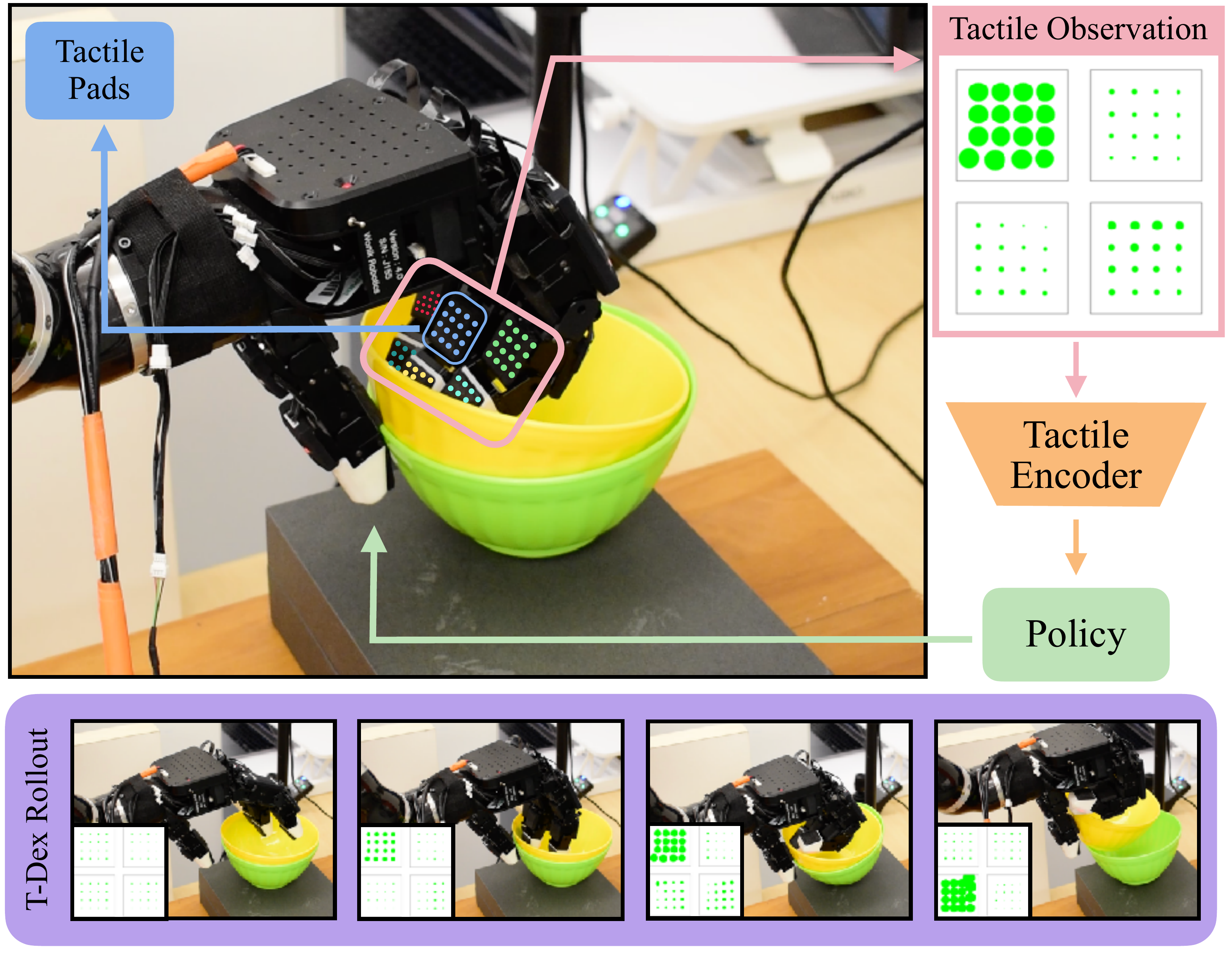}
    \caption{\method{} learns dexterous policies from high-dimensional tactile sensors on a multi-fingered robot hand (top). Combined with vision, our tactile representations are crucial to learn fine-grained manipulation tasks (bottom).}
    \label{fig:figure_1}
\end{figure}

The importance of touch is most evident in human dexterity. Touch is the first sense developed in babies, at as little as 7 weeks in utero~\cite{Fagard2018} and newborn babies have grasping reflexes when touched~\cite{Schott2003}. Moreover, adult humans who have the sense of touch on their hands artificially disabled have reduced dexterity~\cite{johansson1992somatosensory}, highlighting the necessity of tactile feedback. Many works have sought to replicate this sensing modality for robotics, with impressive successes in object identification, pose estimation, and policy learning. Simultaneously, the cost of such sensors is becoming increasingly affordable. However, the application of tactile sensing to multi-fingered dexterous manipulation remains limited.

So, why is tactile-based dexterity hard to achieve? There are three significant challenges. First, tactile sensors are difficult to simulate, which limits the applicability of sim2real based methods~\cite{openai2019learning, wang2021elastic}. Second, for many commonly available tactile sensors, precise calibration of analog readings to physical forces is difficult to achieve~\cite{lee2020calibrating}. This limits the applicability of physics-based control. Third, for multi-fingered hands, tactile sensors need to cover a larger area compared to two-fingered hands. This increases the dimensionality of the tactile observation, which in turn makes learning-based approaches inefficient. A common approach to alleviate this challenge in vision-based learning is to use pretrained models that encode high-dimensional images to low-dimensional representation. However, such pretrained models do not exist for tactile data.

In this work, we present \method{}, a new approach to teach tactile-based dexterous skills on multi-fingered robot hands. To overcome issues in simulating tactile sensors and calibration, we use an imitation framework that trains directly using raw tactile data obtained from a human operator teleoperating the robot. However, directly reasoning about actions from raw tactile data would still require collecting large amounts of demonstrations. To address this, we take inspiration from recent works in robot play~\cite{young2021playful}, and pretrain our own tactile representations. This is done by collecting 2.5 hours of aimless manipulation of objects through teleoperation. Tactile data collected through this play is used to train tactile encoders through standard self-supervised techniques, mitigating the need for exact force calibration.

Given this pretrained tactile encoder, we use it to solve tactile-rich dexterous tasks with just a handful of demonstrations: 6 demonstrations per task, corresponding to under 10 minutes of demonstration time. To achieve imitation with so few demonstrations, we employ a non-parameteric policy that retrieves nearest-neighbor actions from the demonstration set. Importantly, this allows us to combine tactile encodings with other sensor modalities such as vision without any additional training or sensor fusion. This ability to combine touch with vision makes \method{} compatible with tasks that require visual sensing for coarse-grained manipulation and tactile sensing for fine-grained manipulation.

We evaluate \method{} across five challenging tasks such as opening a book, bottle cap opening, and precisely unstacking cups. Through a large-scale experimental study of over 40 hrs of robot evaluation we present the following insights:
\begin{enumerate}
    \item \method{} improves upon vision-only and torque-only imitation models with over a 170\% improvement in average success rate (Section~\ref{sec:tdex_efficiency}).
    \item Play data significantly improves tactile-based imitation, with an average of 58\% improvement over tactile models that do not use play data (Section~\ref{sec:play}).
    \item Ablations on different tactile representations and architectures show that the design decisions in \method{} are important for high performance (Section~\ref{sec:arch}).
\end{enumerate}
Open-sourced code, data and videos of \method{} can be found at \website{}.

\section{Related Work}

Our work builds on several prior ideas in dexterous manipulation, tactile sensing, representation learning and imitation learning. For brevity, we describe the most relevant below: 

\subsection{Dexterous Manipulation}
Multi-fingered robot control has been studied extensively~\cite{Ciocarlie2007DexterousGV, kumar2014real, Shigemi2018}. Initial work focuses on physics-based modelling of grasping \cite{okamura2000overview, odhner2014compliant} that often used contact force estimates to compute grasp stability. However, contact estimates derived from motor torque only give point estimates and are susceptible to noise due to coupling with the hand's controller. There has also been work on performing dynamic tasks with tactile sensing~\cite{ishihara2006dynamic} that relies on a hand-designed controller. More recent works have sought to incorporate learning into the process to reduce the need for accurate modeling and to allow for more complicated manipulation tasks. 

There are several methodologies for using learning in dexterity. Model-based reinforcement learning (RL) methods have been shown to work in both simulation~\cite{mordatch2012contact, lowrey2018plan} and the real world~\cite{Nagabandi2019, kumar2016dext}.
Model-free RL has been used to train policies both in simulation~\cite{huang2021generalization, chen2021system} and directly on hardware~\cite{Zhu2019}. Simulation to real transfer has also shown success~\cite{lowrey2018reinforcement, openai2019learning, nvidia2022dextreme, 9812093}, though it often requires extensive randomization, significantly increasing training time. The use of expert demonstrations can reduce the amount of real-world interactions needed to learn a dexterous policy~\cite{Rajeswaran2018, Zhu2019, arunachalam2022dexterous}. The works mentioned above either use visual observations or estimates of object state, which suffer during heavy occlusion of the object.

\subsection{Tactile Sensing}

To give robots a sense of touch, many tactile sensors have been created for enhancing robotic sensing~\cite{https://doi.org/10.48550/arxiv.2106.08851, bhirangi2021reskin, alspach2019soft}. Prominently, the GelSight sensor has been used for object identification~\cite{patel2021digger}, geometry sensing~\cite{dong2017improved}, and pose estimation~\cite{kelestemur2022tactile}. However, since GelSight requires a large form factor, it is difficult to cover a entire multifingered hand with it. Instead `skin'-like sensors~\cite{8858052} and tactile pads can cover entire hands, yielding high-dimensional tactile observations for dexterity. In this work, we use the XELA uSkin \cite{8307485} sensors to cover our Allegro hand.

Due to the high-dimensional readings from tactile sensors, machine learning has been employed to leverage the sensors for a variety of applications. The sensors have been applied to two-fingered grippers to improve grasping and manipulation~\cite{10.1007/978-3-030-33950-0_33, calandra2018more, https://doi.org/10.48550/arxiv.1910.02860}. They have also been used with cameras for object classification~\cite{zambelli2021learning}, 3D shape detection~\cite{wang20183d}, and to learn a multi-modal representation for end effector control~\cite{https://doi.org/10.48550/arxiv.2209.13042, https://doi.org/10.48550/arxiv.1907.13098}.
However, these prior works differ from \method{} in two key ways. First, such tactile learning methods have not been applied to multifingered hands. Second, the tactile representations learned in these works require large amounts of task-centric data for each task. On the other hand, \method{} uses a large amount of task-agnostic play data, which enables learning tasks with small amounts of data per task. 

\subsection{Representation Learning for Robotics}

Learning concise representations from high-dimensional observations is an active area of research in robotics. A wide variety of approaches using auto-encoders~\cite{finn2016deep, MVP, ha2018world}, physical interaction~\cite{Pinto2016}, dense descriptors~\cite{florence2018dense}, and mid-level features~\cite{chen2020robust} have been studied.

In computer vision, self-supervised learning (SSL) is often used to pre-train visual features from unlabeled data, improving downstream task performance. 
Contrastive methods learn features by moving features of similar observations closer to one another and features of dissimilar observations farther from one another~\cite{simclr, caron2020unsupervised}. These methods require sampling negative pairs of datapoints, which adds an additional layer of complexity. Non-contrastive methods typically try to learn features by making augmented versions of the same observation close~\cite{grill2020bootstrap, bardes2021vicreg} and do not require sampling negative examples. 
Self-supervision has been adopted for visual RL~\cite{yarats2021reinforcement, laskin2020curl, nair2022r3m, MVP} and robotics~\cite{sermanet2018time, pari2021surprising, holodex, zhan2020framework} to improve sample efficiency and asymptotic performance. SSL methods have also been applied to other sensory inputs like audio~\cite{niizumi2023byol-a} and depth~\cite{afham2022crosspoint}.
We build on this idea of self-supervision and extend it to tactile observations. However, unlike visual data, for which large pretrained models or Internet data exists, neither are available for tactile data. This necessitates the creation of large tactile datasets, which we generate through robot play.

\subsection{Exploratory and Play Data}
Since task-specific data can be expensive to collect, a number of works have examined leveraging off-policy data to improve task performance. Previous work has used play data to learn latent plan representations~\cite{lynch2019learning} and to learn a goal-conditioned policy~\cite{cui2022play}. 
Recent work in offline RL has noted that including exploratory data improves downstream performance~\cite{yarats2022don} and that actively straying away from the task improves robustness~\cite{https://doi.org/10.48550/arxiv.2210.02343}. These findings are paralleled by studies on motor development in humans. 3-5-month old infants spontaneously explore novel objects~\cite{rochat1989object} and 15-month-old infants produce the same quantity of locomotion in a room without toys than in a room with toys~\cite{Hoch2018}. Given these motivating factors, we opt to leverage a play dataset of cheap, imperfect, tactile-rich interactions in order to improve our representations and downstream task performance.

\subsection{Offline Imitation Learning}
Imitation Learning (IL) allows for efficient training of skills from from data demonstrated by an expert. Given a set of demonstrations, offline imitation methods such as Behavior Cloning (BC) use supervised learning to learn a policy that outputs actions similar to the expert data and have been used extensively in robotics~\cite{pomerleau1989alvinn, florence2022implicit, shafiullah2022behavior, zhu2022viola, mandlekar2021matters}. However, such methods often require demonstrations on the order of hundreds to thousands trajectories. While getting such a scale of demonstrations for two-fingered grippers is achievable using a surrogate for the robot hardware~\cite{song2020grasping, young2020visual}, collecting the same quantity of data for dexterous tasks is difficult due to cognitive and physical demands of teleoperating multi-fingered hands. To learn with fewer demonstrations in high-dimensional action spaces, non-parametric approaches such as nearest neighbors have shown to be more effective than parametric ones~\cite{arunachalam2022dexterous, holodex}. \method{} builds on this idea and uses nearest neighbor-based offline imitation to learn tasks with few demonstrations.

\begin{figure}[ht]
    \centering
    \includegraphics[width=\linewidth]{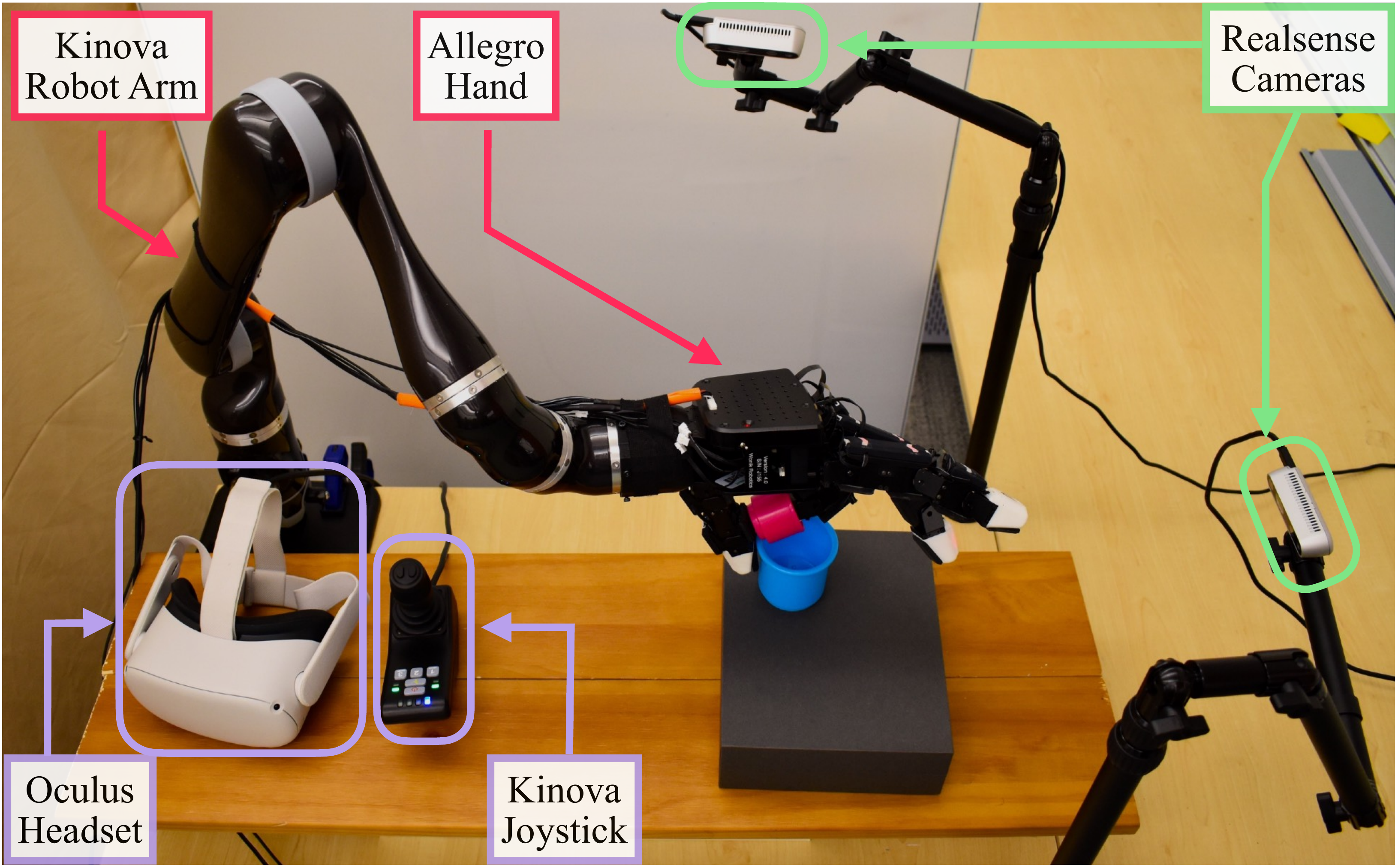}
    \caption{Hardware setting of \method{}. We use an Oculus Headset to teleoperate the Allegro hand and the built in Kinova joystick to control the arm. Visual observations are streamed through two different Realsense cameras and tactile observations are saved with XELA touch sensors on the Allegro hand.}
    \label{fig:robot_setup}
\end{figure}

\section{System Details and Robot Setup}
\label{sec:system}

Our robotic system, visualized in Figure \ref{fig:robot_setup}, consists of a robotic arm and hand. The arm is a 6-dof Kinova Jaco and the hand is a 16-dof Allegro hand with four fingers. The arm can be teleoperated through the built-in Kinova joystick, while the the hand can be teleoperated using the the Holo-Dex framework~\cite{holodex}. Here, our teleoperator uses a virtual reality headset to both visualize robot images and control the hand in real time. The headset returns a pose estimate for each finger of the hand which is re-targeted to the Allegro Hand. Inverse Kinematics is then used to translate target Cartesian positions in space to joint angles, which are fed into the low-level hand controller. To achieve robust position control, we use a low-level PD joint position controller with gravity compensation to allow the robot to maintain a hand pose at different orientations in space. Our action space is Cartesian position and orientation of the arm (3D position and 4D quaternion for orientation) and the 16-dimensional joint state of the hand for a total of 23 dimensions.

The Allegro hand is fitted with 15 XELA uSkin tactile sensors, 4 on each finger and 3 on the thumb. Each sensor has a 4x4 resolution output of tri-axial force reading (forces in translational x, y, z) information, which amounts to a 720-dimensional tactile reading. The force readings are uncalibrated, susceptible to hysterisis, and can change when strong magnets or metals are in the vicinity. Due to this, we opt against explicit calibration of the 720 sensor units. To supplement the tactile sensors, we also use two RGB cameras with 640x480 resolution to capture visual information in the scene, though our policies only uses information from one to execute. Our choice of camera for tasks depends on which one captures the most visual information about the objects to ensure fairness when comparing to baselines and enable better joint vision and tactile control.

\section{Tactile-Based Dexterity (\method{})}

\begin{figure*}[t]
    \includegraphics[width=\textwidth]{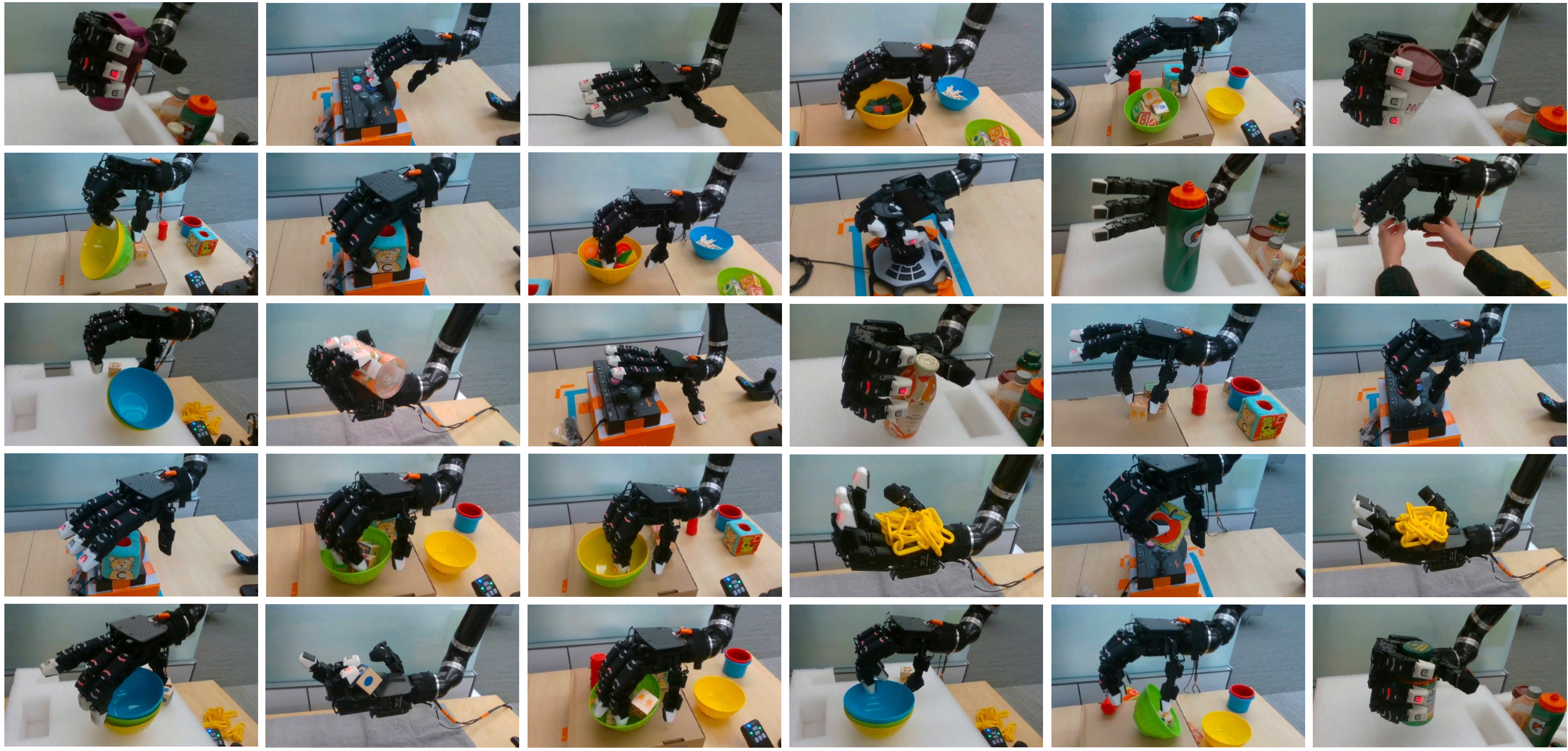}
    \caption{Visualization of some of the play tasks.  We play with grasping, pinching, moving objects, and other in-hand manipulation tasks.}
    \label{fig:play_tasks}
\end{figure*}

\method{} operates in two phases: pretraining from task-agnostic play data and downstream learning from a few task-specific demonstrations. In the pretraining phase, we begin by collecting a diverse, contact-rich play dataset from a variety of objects by teleoperating the robot (see Section~\ref{sec:system} for details). Once collected, we use self-supervised learning (SSL) algorithms on the play data to learn an encoder for tactile observations. In the downstream learning phase, a teleoperator collects demonstrations of solving a desired task. Given these demonstrations, non-parametric imitation learning is combined with the pretrained tactile encoder to efficiently learn dexterous policies. See Figure \ref{fig:ssl_imitation} for a high-level overview of our framework.
Details of individual phases follow:

\subsection{Phase I: Pre-Training Tactile Representations from Play}
\label{sec:phase_1}
\textbf{Play Data Collection:}
The play data is collected from a variety of contact-rich tasks including picking up objects, grasping a steering wheel, and in-hand manipulation. Visualization of some of the play tasks can be seen in Figure \ref{fig:play_tasks}. We collect a total of 2.5 hours of play data, including failed examples and random behavior.
Because the image and tactile sensors operate at 30Hz and 100Hz, respectively, we sub-sample the data to about 10Hz to reduce the size of the dataset. We only include observations whenever the total changed distance of the fingertips and robot end effector exceed 1cm, reducing the dataset from 450k frames to 42k. This reduces subsequent training time and filters out similar states in the dataset when the robot is still, which could potentially bias the SSL phase. All of the play data will be publicly released on our website at \website{}.

\textbf{Feature Learning:}
To extract useful representations from the play data we employ SSL, which tries to learn a low dimensional representation from high-dimensional observations~\cite{ericsson2022self}. Specifically, we use Bootstrap your own Latent (BYOL), which has been shown to improve performance on computer vision tasks~\cite{grill2020bootstrap} as well as visual robotics tasks~\cite{pari2021surprising, arunachalam2022dexterous}. 
BYOL has both a primary encoder $f_\theta$, and a target encoder $f_\xi$, which is an exponential moving average of the primary. Two augmented views of the same observation  $o$ and $o'$ are fed into each to produce representations $y$ and $y'$, which are passed through projectors $g_\theta$ and $g_\xi$ to produce $z$ and $z'$, which are higher dimensional. The primary encoder and projector are then tasked with predicting the output of the target projector. After training, we use $f_\theta$ to extract features from observations.

\begin{figure*}[t]
    \centering
    \includegraphics[width=\textwidth]{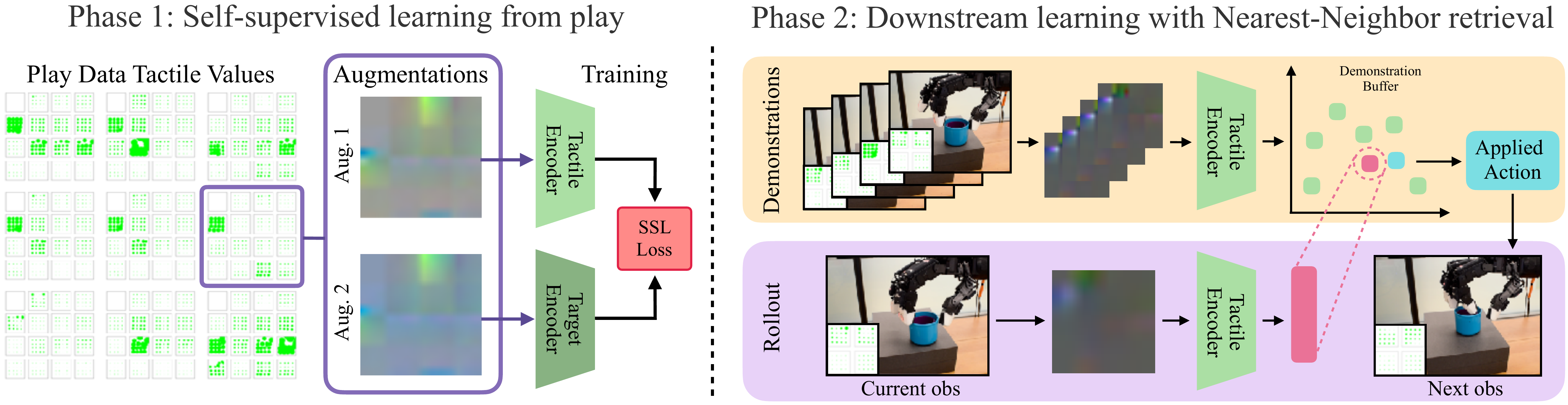}
    \caption{An overview of the \method{} framework. Left: we train tactile representations using BYOL on a large play dataset. Right: we leverage the learned representations using nearest neighbors imitation.}
    \label{fig:ssl_imitation}
\end{figure*}

To apply BYOL to our tactile data, we treat the sensor data as an image with one channel for each axis of force. Each of the finger's 3-axis 4x4 sensors are stacked into a column to produce a 16x4 image for the fingers and a 12x4 image for the thumb. These images are then concatenated to produce a three-channel 16x16 image with constant padding for the shorter thumb. A visualization of the tactile images can be seen in Figure \ref{fig:ssl_imitation}. We scale the tactile image up to 224x224 to work with standard image encoders. For the majority of our experiments, we use the AlexNet~\cite{krizhevsky2012imagenet} architecture, also starting with pre-trained weights.
Unlike SSL techniques in vision~\cite{grill2020bootstrap}, we only apply the Gaussian blur and small random resized crop augmentations, since other augmentations such as color jitter and grayscale would violate the assumption that augmentations do not change the tactile signal significantly. Importantly, unlike vision, since all of the tactile data is collected in the frame of the hand, the sensor readings are invariant to hand pose and can be easily reused between tasks.

\subsection{Phase II: Non-parametric Learning}
\textbf{Demonstration Collection:}
Six demonstrations are collected for each task by a teleoperator. Because the nature of our tasks are contact-dependent and the human operator does not receive tactile feedback, there is a relatively high failure rate while collecting demonstrations.
Although the successful demonstrations correspond to at most 10 minutes of robot time, it requires up to 30 minutes of collection time in order to get successful demonstrations. A common complaint for our teleoperators was that it is difficult to infer tactile feedback in Holodex~\cite{holodex}, which resulted in a large fraction of failures. This exemplifies why tactile feedback is necessary to accelerate learning of dexterous manipulation and the importance of learning from small amounts of task-specific data. 

Similar to the play data, we subsample the demonstrations to only include data where the fingertips and end effector of the arm move by more than a total of 2cm. This can be viewed as discretizing space rather than time, reducing the size of the dataset and serving as a filtering mechanism to remove noise in the demonstrations.

\textbf{Visual Feature Learning:}
Many of our tasks require coarse-grained information about the location of objects. This necessitates incorporating vision feedback as tactile observations are not meaningful when the hand is not touching the object. To do this, we extract visual features using standard BYOL augmentations on the images collected from demonstration data. The views for each task are significantly different, so we did not observe a benefit from including the play data in the visual representation learning. Similar to prior work~\cite{arunachalam2022dexterous, holodex}, we start with a ResNet-18~\cite{he2016deep} architecture that has been pre-trained on the ImageNet~\cite{deng2009imagenet} classification task. 

\textbf{Downstream Imitation Learning:}
Our action space consists of both the hand pose, specified by 16 absolute joint angles, and the robot end effector position and orientation, specified by a 3-dimensional position and a 4-dimensional quaternion. Due to both the high-dimensional action and observation spaces, parametric methods struggle to learn quality policies in the low-data regime. To mitigate this, we use a nearest neighbors-based imitation learning policy~\cite{pari2021surprising} to leverage our demonstrations. For each tuple of observations and actions in the demonstrations $(o_i^V, o_i^T, a_i)$, we compute visual and tactile features $(y_i^V, y_i^T)$ and store them along-side the corresponding actions. Since the scales of the two features are at different, we scale both features such that the maximum distance in the dataset for each feature is 1. At test time $t$ given $o_t$, we compute $(y_t^V, y_t^T)$, find the datum with the lowest total distance, and execute the action associated with it.

\section{Experiments}

We evaluate \method{} on a range of tasks that are designed to answer the following questions:
\begin{itemize}
    \item Does tactile information improve policy performance?
    \item How important is play data to our representations?
    \item What are important design choices when learning features of tactile information?
\end{itemize}

\begin{figure*}[t]
    \centering
    \includegraphics[width=\textwidth]{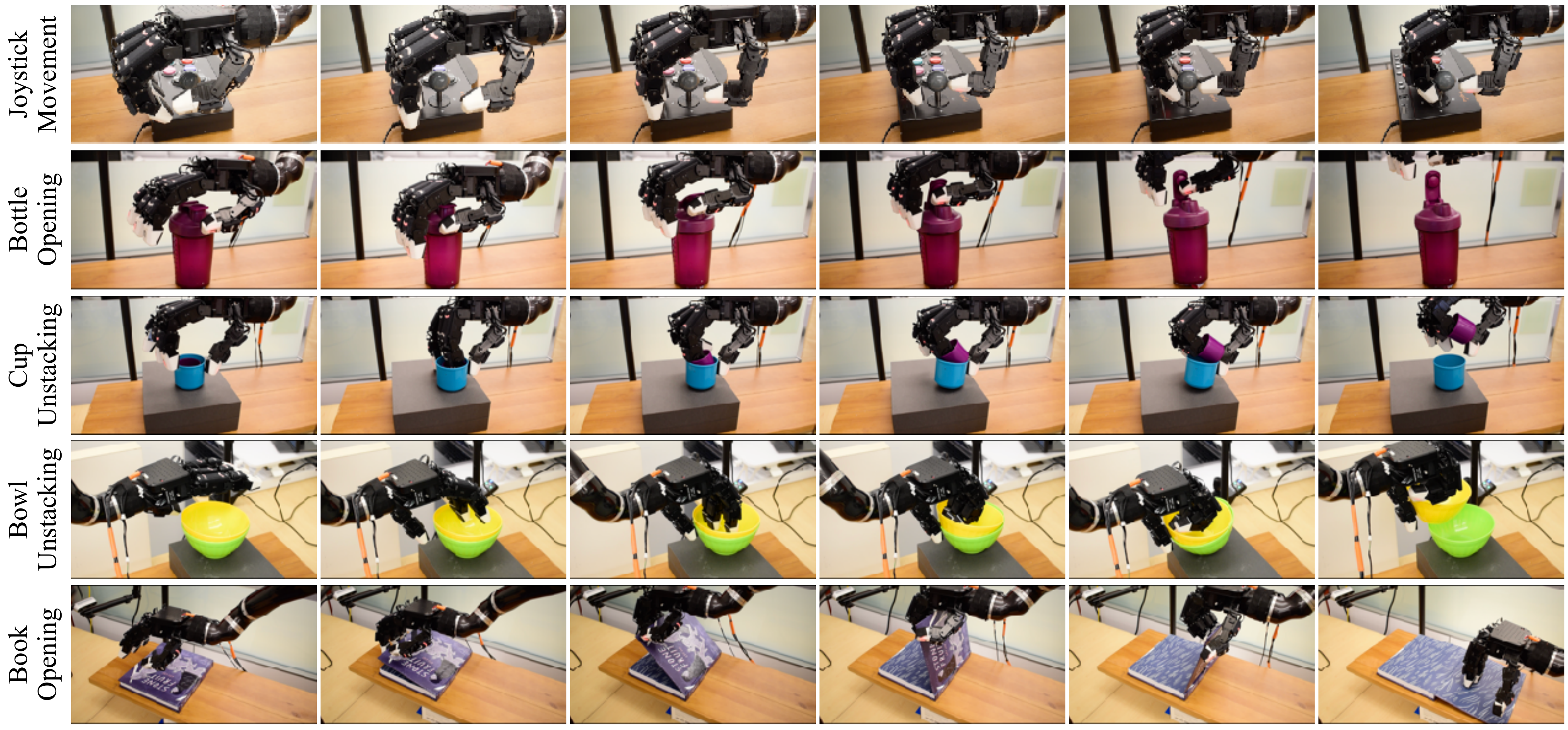}
    \caption{Visualization of robot rollouts from \method{} policies. Note the severe visual occlusions when the robot makes contact with the object.}
    \label{fig:policy_rollout}
        \vspace{-0.2cm}
\end{figure*}

\subsection{Description of Dexterous Tasks}
We examine five dexterous contact-rich tasks that require precise multi-finger control (see Figure \ref{fig:policy_rollout}). We describe them in detail below: 

\subsubsection{Joystick Movement} Starting over an arcade gamepad, the hand is tasked with moving down and pulling a joystick backwards. This task is difficult because the hand occludes the gamepad when manipulating it. We collect demonstrations of the joystick in two different positions and evaluate on different positions and orientations not seen during training. A trial is successful if the joystick has been pulled within 60 seconds.
\subsubsection{Bottle Opening} This task requires the hand to open the lid of a bottle. We collect three demonstrations with the bottle orientation requiring the use of the thumb, and three other requiring the use of the middle finger. The task is considered successful if the lid is open within 120 seconds.
\subsubsection{Cup Unstacking} Given two cups stacked inside one another, the tasks is to remove the smaller cup from the inside of the larger one. In addition to occlusion, this task requires making contact both the inner and outer cups before lifting the inner cup with the index finger. It is considered a success if the smaller cup is raised outside the larger cup without dropping it or knocking the cup off the table within 240 seconds.
\subsubsection{Bowl Unstacking} This task is similar to the previous, but with bowls instead of cups. Since the bowls are larger, multiple fingers are required to lift and stabilize them. A run is successful if it has lifted the bowl within 100 seconds.
\subsubsection{Book Opening} This task requires opening a book with three fingers. After making contact with the cover, the hand must pull up with an arm movement, remaining in contact until it is fully open. The task is considered a success if the book is open within 300 seconds.

To evaluate various models for dexterity, we first collect six demonstrations for each task in which the object's configuration is varying inside a 10x15cm box. Models are then evaluated on new configurations in the convex hull of demonstrated ones. This follows the standard practice of evaluating representations for robotics~\cite{zhou2022train, nair2022r3m, MVP}. Additional experimental details can be found in Appendix \ref{sec:exp_details}.

\subsection{Baselines for Dexterity}
We study the impact of tactile information on policies learned through imitation, comparing against a number of baselines. Unless otherwise specified, the methods receive both tactile and image data. Each method is described below:

\subsubsection[Behavior Cloning (BC)]{Behavior Cloning (BC)~\cite{Pomerleau1989}} We train a neural network end-to-end to map from visual and tactile features to actions.

\subsubsection[Nearest Neighbors with Torque only (NN-Torque)]{Nearest Neighbors with Torque only (NN-Torque)~\cite{9812093}} We perform nearest neighbors with the output torques from our PD controller. The torque targets can be used as a proxy for force, providing some tactile information.

\subsubsection[Nearest Neighbors with Image only (NN-Image)]{Nearest Neighbors with Image only (NN-Image)~\cite{pari2021surprising}} We perform nearest neighbors with the image features only. During evaluation, to ensure fairness we use viewpoints that can convey maximal information about the scene.

\subsubsection{Nearest Neighbors with Tactile only (NN-Tactile)} Nearest neighbors with the tactile features trained on play data. Unlike \method{} we do not use vision data for this baseline.

\subsubsection{Nearest Neighbors with Tactile Trained on Task Data (NN-Task)} Instead of training the tactile encoder on the play data, we train it on the 6 task-specific demonstrations.

\subsubsection{Nearest Neighbors with Tactile Trained on Play Data (\method{})} This is our main method with the tactile encoder pre-trained on all the play data followed by nearest neighbor retrieval on task data.

Additional model details can be found in Appendix \ref{sec:model_details}. 
\subsection{How important is tactile sensing for dexterity?}

\begin{figure*}
    \centering
    \includegraphics[width=\textwidth]{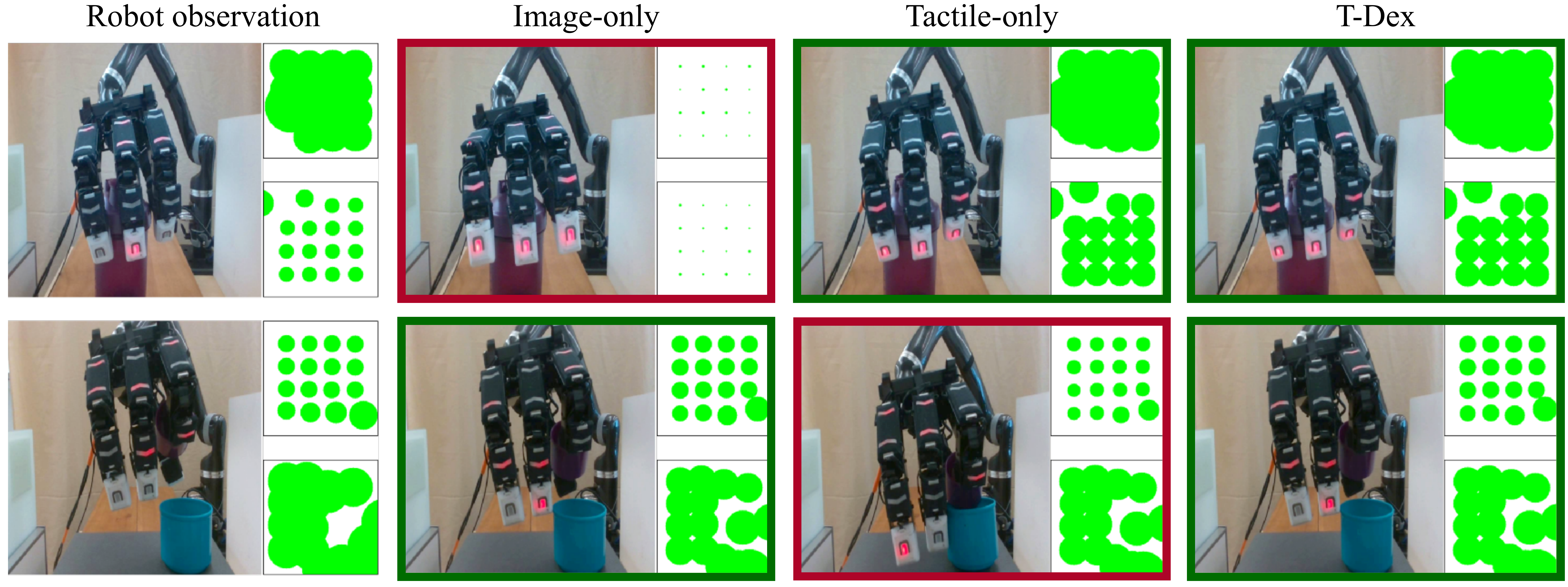}
    \caption{Visualization of the camera image, top-two activated tactile sensors, and their nearest neighbors for our method and baselines. While it is able to return a visually similar image, the image-only baseline is unable to recognize contact with the bottle. The tactile-only baseline can return a tactilely similar neighbor, but fails to capture the position of the robot.}
    \label{fig:nn}
\end{figure*}

\begin{table}[t]
\centering
    \caption{Real-world success rate of the learned policies}

    \begin{tabular}{c|ccccc|c}
               & Joystick & Cup & Bowl & Book & Bottle & Average \\ \hline
    BC         & 0\%     & 0\%         & 0\%          & 0\%          & 0\%            & 0\%     \\
    NN-Image   & 40\%    & 0\%         & 20\%         & 50\%         & 0\%            & 22\%    \\
    NN-Tactile & 60\%    & 0\%         & 20\%         & 0\%          & \textbf{60\%}           & 28\%    \\
        NN-Task    & \textbf{80\%}    & 40\%        & 30\%         & 60\%         & 30\%           & 48\%    \\
    NN-Torque & 70\% & 20\% & 40\% & 30\% &30\% & 38\% \\
    \textbf{\method}     & \textbf{80\%}    & \textbf{80\%}        & \textbf{70\%}         & \textbf{90\%}         & \textbf{60\%}           & \textbf{76\%}   
    \end{tabular}
    \label{tab:success}
    \vspace{-0.2cm}
\end{table}

\label{sec:tdex_efficiency}
In Table \ref{tab:success} we report success rates of \method{} along with baseline methods. We make several observations from these experiments. First, we find that BC completely fails on all tasks, quickly moving to states outside the distribution of demonstrations. This behavior has been previously reported in small data regime~\cite{arunachalam2022dexterous}. Among the nearest neighbor based methods, we find that tactile-only (NN-tactile) struggles on Book Opening and Cup Unstacking since the hand fails to localize the objects to make first contact. On the other hand, the image-only (NN-Image) struggles on Bottle Opening and Cup Unstacking as severe occlusions caused by the hand result in poor retrievals. Using torque targets (NN-Torque) instead of tactile feedback proved useful, improving over NN-Image, but did not match using tactile feedback.

\begin{figure}[t]
    \centering
    \includegraphics[width=0.95\linewidth]{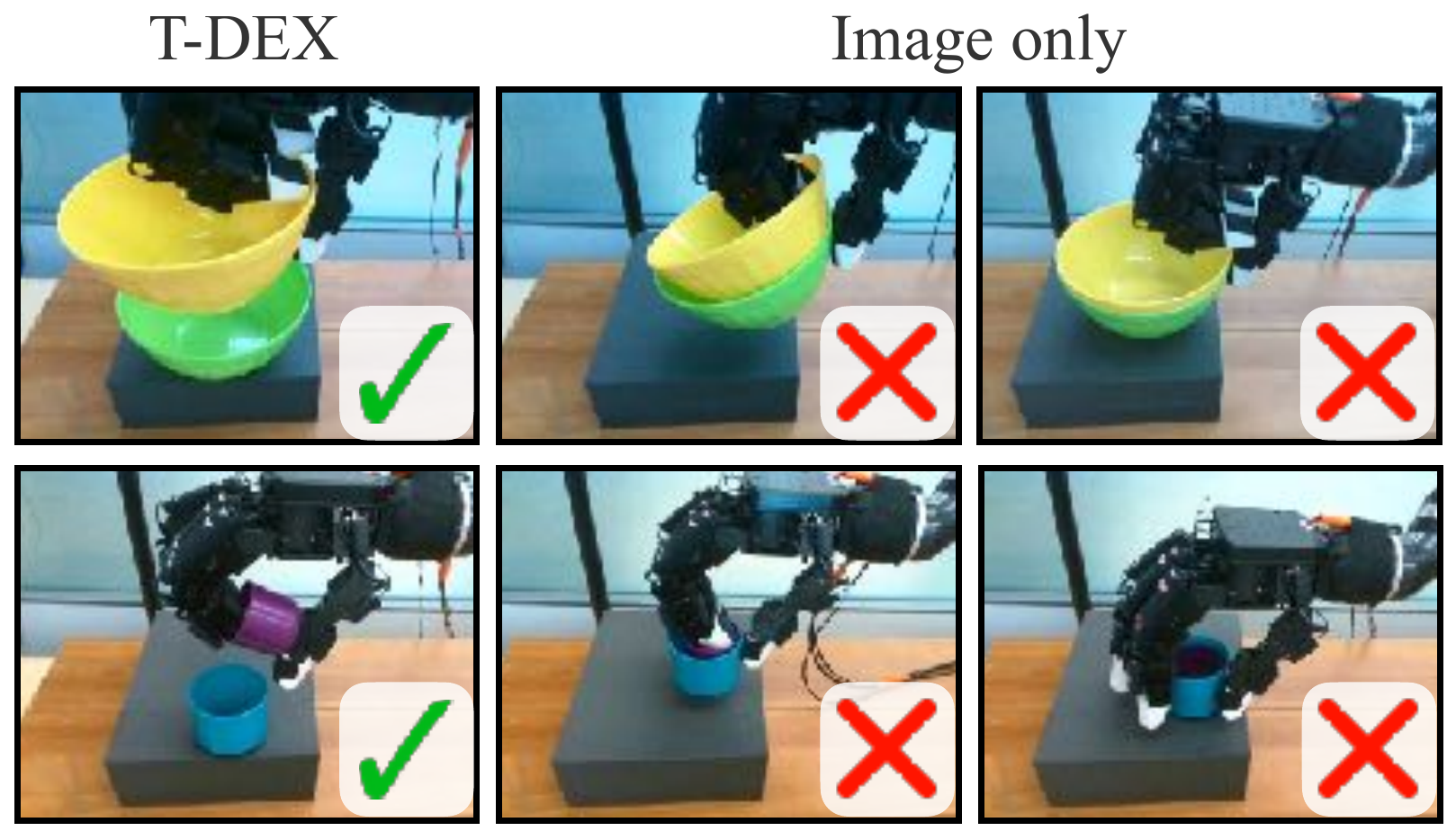}
    \caption{Visualization of the failure modes of our Image only baseline. Without tactile information, the robot applies either too much force, causing it to pick up both objects or does not realize it is not correctly in contact with the objects.}
    \label{fig:failures}
    \vspace{-0.2cm}
\end{figure}

We find that \method{} combines the coarse-grained localization ability of NN-Image along with the fine-grained manipulation of NN-Tactile, and results in the strongest results across all tasks. To further analyze why \method{} performs so well, we visualize the nearest neighbors of states for the image-only and tactile-only methods. Figure \ref{fig:nn} shows neighbors for our method and baselines. Our method produces neighbors that seem to capture the state of the world better than image and tactile alone. On the Bottle Opening task, the NN-Image method returned a neighbor that looked visually similar, but was not in contact with the bottle. The NN-Tactile method was able to produce a similar grasp to the observation, but it was not able to capture the position of the hand. Additional failure modes for NN-Image can be see in Figure \ref{fig:failures}. We notice that it often applies too much or not enough force, causing the task to fail. Combining both image and tactile information gives us the best of both, allowing us to find a visually and tactilely similar neighbor. Successful policy rollouts can be seen in Figure \ref{fig:policy_rollout} and in Appendix \ref{sec:ap_rollouts}.

\subsection{Does pre-training on play improve tactile representations?}
\label{sec:play}

To understand the importance of pre-training, we run NN-Task, which pre-trains tactile representations on task data. As seen in Table \ref{tab:success}, This baseline does quite well on the simpler Joystick Movement task. However, on harder tasks, particularly the Unstacking tasks and Bottle Opening, we find that NN-Task struggles significantly. This can be attributed to poor representational matches when trained on limited task data. To mitigate this, we also try training the encoder with a combination of successful and failed demonstrations on the Bowl Unstacking task, getting a success rate of 30\%, which shows no improvement in task performance. 

\begin{figure}
    \centering
    \includegraphics[width=0.5\textwidth]{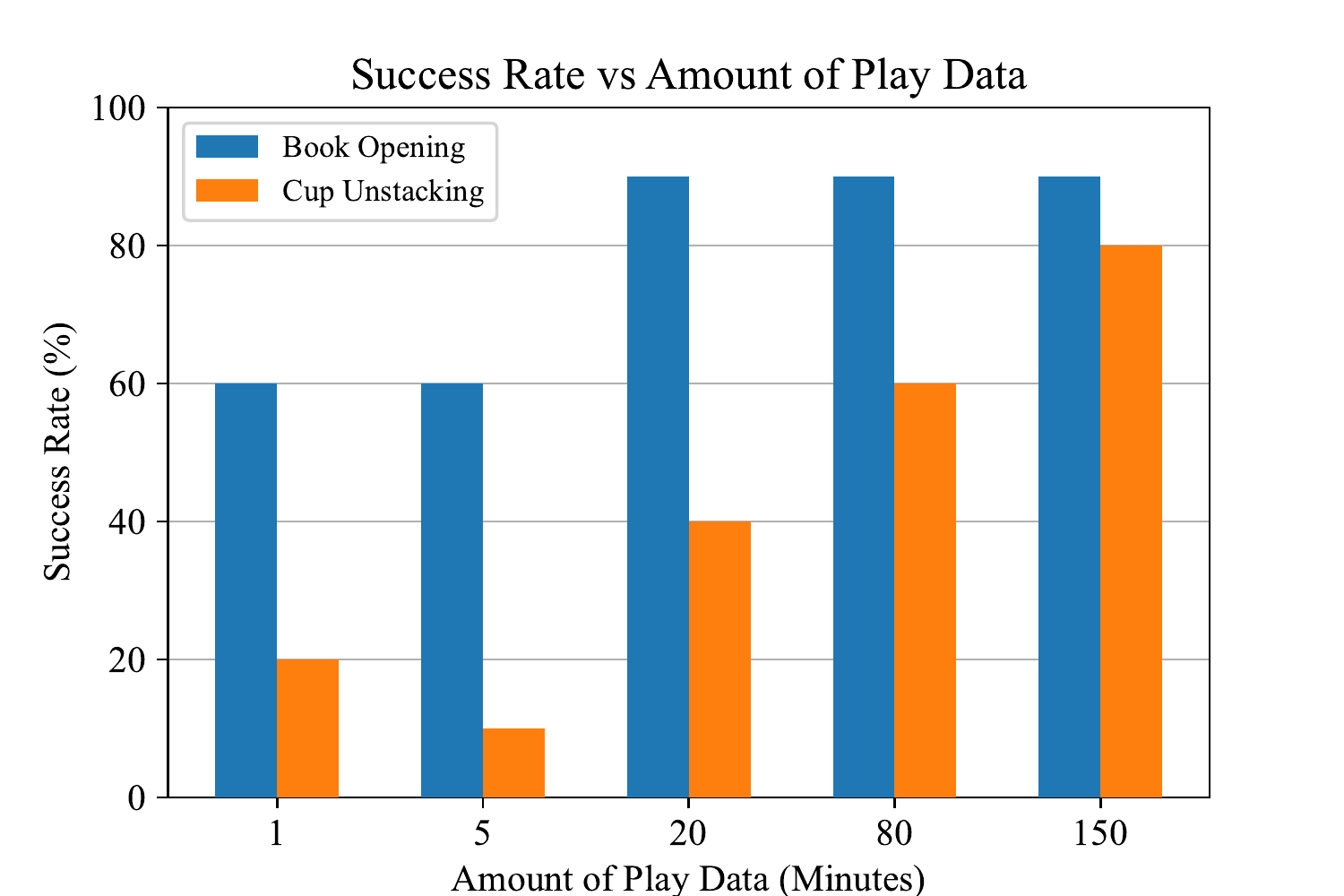}
    \caption{Success rate on Book Opening and Cup Unstacking tasks with varying amount of play data. Training only on task data performs moderately well, but is outperformed with just 20 minutes of play.}
    \label{fig:play_data_amt}
    \vspace{-0.5cm}
\end{figure}

To provide further evidence for the usefulness of tactile pretraining, we plot the gains in performance across varying amounts of play data in Figure~\ref{fig:play_data_amt}. We see that for easier tasks like Book Opening, even small amounts of play data (20 mins) is sufficient to achieve a 90\% success rate. However, for harder tasks like Cup Unstacking, we see steady improvements in success rate with larger amounts of play data. 

\subsection{Importance of tactile representation}
\label{sec:arch}

\begin{figure*}[t]
    \centering
    \includegraphics[width=\textwidth]{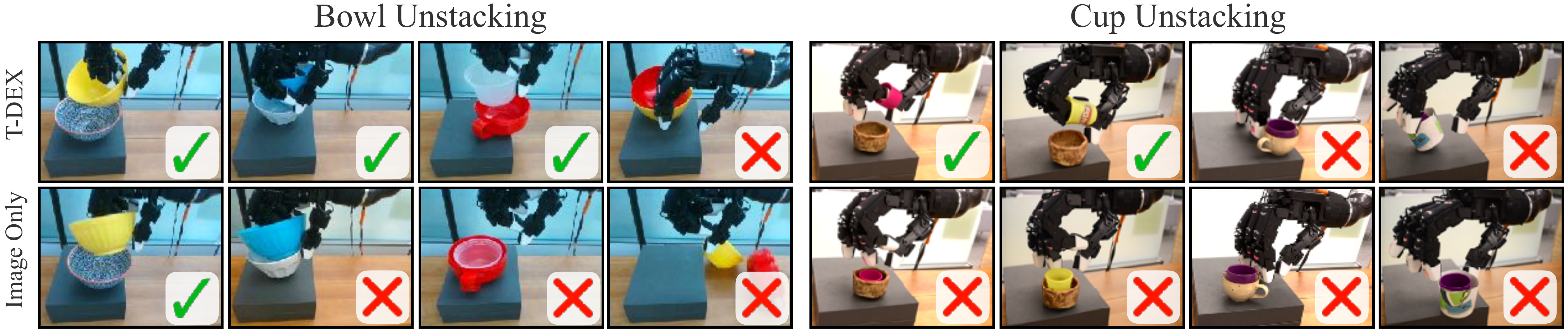}
    \caption{We show rollouts of \method{} and our Image-only baseline on objects not seen during demonstration collection. For both Bowl and Cup Unstacking, we find that \method{} generalizes to more scenarios, while our image only baseline fails on all but one scenario.}
    \label{fig:gen}
\end{figure*}

A critical component in \method{} is the architectural details in representing and processing tactile data. In this section, we examine various architectures to represent tactile features. For simplicity, we study a subset of the tasks, Book Opening and Cup Unstacking. Each encoder is trained using BYOL on the play dataset with the same augmentations used in the main method. We compare our main encoder, AlexNet with different architectures described below:
\begin{itemize}
    \item \textit{ResNet:} A standard ResNet-18~\cite{he2016deep} with weights pre-trained on the ImageNet~\cite{deng2009imagenet} classification task.
    \item \textit{3-layer CNN:} A lightweight CNN with three layers initialized with random weights.
    \item \textit{Stacked CNN:} Rather than laying out the sensor data of the fingers spatially in the image, we consider stacking the sensor output into one 45-channel image.
    \item \textit{Shared CNN:} We consider a shared encoder for each sensor pad. We pass individual pad values to the same network and concatenate the outputs.
    \item \textit{Raw Tactile:} Instead of utilizing the geometry of the tactile sensors with a CNN, we flatten the raw tactile data into a 720-dimensional vector.
\end{itemize}

\begin{table}[t]
    \centering
    \caption{Success rates of various representations for tactile data on the Book Opening and Cup Unstacking tasks.}

    \begin{tabular}{c|cccccc}
             & \method & ResNet & 3-layer & Stacked & Shared & Raw \\ \hline
Book  & \textbf{90\%}           & \textbf{90\%}   & 60\%        & 50\%        & 20\%       & 50\%        \\
Cup  & \textbf{80\%}           & 60\%   & 30\%        & 10\%        & 10\%       & 30\%       
\end{tabular}
    \label{tab:arch}
    \vspace{-0.3cm}
\end{table}

Full results for this experiment can be found in Table \ref{tab:arch}. We find that both \method{} and ResNet perform similarly on Book Opening, although ResNet takes significantly more computation for the same results. On Cup Unstacking we find that ResNet performs a little worse than \method{}, which further informs our architectural choice. While, one may conclude that smaller architectures are better, we see that a simpler 3-layered CNN also performs poorly and does not reach the performance of either of the larger models.

Apart from the architecture, we find that the structure of inputting tactile data from individual tactile pads is also important. For example, we find that stacking tactile pads channel-wise is substantially worse than \method{} that stacks the tactile pads spatially. Similarly we find that using a shared encoder for each tactile pad is also poor. This is perhaps because of the noise that exists in high-dimensional raw tactile data, which is difficult to filter out with the stacked and shared encoders. Hence, one spurious reading in an unused tactile pad could yield an incorrect neighbor, producing a bad action. This hypothesis is further substantiated in the Raw Tactile method, which is roughly on par with the Stacked method.

We additionally run three experiments with different tactile representations on the Bowl Unstacking task to analyze our choice of representation. We run PCA on the Raw Tactile features on the play dataset and use the top 100 components as features, achieving a success rate of 40\%. When PCA fails, it is not able to capture fine-grained tactile information that is necessary to solve the task.
Next, we sum the activations of each 4x4 tactile sensor in each dimension to create a 45-dimensional feature, which does not succeed on any task. Finally, we shuffle the order of the pads in the tactile image, which achieves 20\% success, which is much lower than using the structured layout (Section \ref{sec:phase_1}),
showing that the layout of tactile data is highly important to our method.

\subsection{Generalization to Unseen Objects}
To examine the generalization ability of \method{}, we run the Bowl Unstacking and Cup Unstacking tasks with unseen crockery and compare against NN-Image. Seen in Figure \ref{fig:gen}, \method{} is able to pick up 3 out of 4 of the new bowls and 2 out of 4 of the unseen cups, while NN-Image only succeeds on a single novel bowl. Although the learned visual features for the task are only sufficient to solve one of the new tasks, incorporating tactile information enables us to find good neighbors in our demonstrations, allowing us to solve tasks on unseen objects without retraining.

\section{Limitations and Conclusion} 
\label{sec:conclusion}
In this work, we have presented an approach for tactile-based dexterity (\method{}) that combines tactile pretraining on play data along with efficient downstream learning on a small amount of task-specific data. Our results indicate that \method{} can significantly improve over prior approaches that use images, torque and tactile data. However, we recognize two key limitations. Although \method{} succeeds on several out-of-distribution examples, the success rate is lower than the training object. The second, is that our approach is currently limited to offline imitation, which limits the ability of our policies to learn from failures. Both limitations could be addressed by integrating online learning, improving architectures for tactile data, and better tactile-vision fusion algorithms. While these aspects are out of scope to this work, we hope that the ideas introduced in \method{} can spur future work in these directions.

\section*{Acknowledgments}
We thank Vaibhav Mathur, Jeff Cui, Ilija Radosavovic, Wenzhen Yuan and Chris Paxton for valuable feedback and discussions. This work was supported by grants from Honda, Meta, Amazon, and ONR awards N00014-21-1-2758 and N00014-22-1-2773.

\bibliographystyle{plainnat}
\bibliography{ref}

\begin{thebibliography}{80}
\providecommand{\natexlab}[1]{#1}
\providecommand{\url}[1]{\texttt{#1}}
\expandafter\ifx\csname urlstyle\endcsname\relax
  \providecommand{\doi}[1]{doi: #1}\else
  \providecommand{\doi}{doi: \begingroup \urlstyle{rm}\Url}\fi

\bibitem[Afham et~al.(2022)Afham, Dissanayake, Dissanayake, Dharmasiri,
  Thilakarathna, and Rodrigo]{afham2022crosspoint}
Mohamed Afham, Isuru Dissanayake, Dinithi Dissanayake, Amaya Dharmasiri,
  Kanchana Thilakarathna, and Ranga Rodrigo.
\newblock Crosspoint: Self-supervised cross-modal contrastive learning for 3d
  point cloud understanding.
\newblock In \emph{Proceedings of the IEEE/CVF Conference on Computer Vision
  and Pattern Recognition}, pages 9902--9912, 2022.

\bibitem[Alspach et~al.(2019)Alspach, Hashimoto, Kuppuswamy, and
  Tedrake]{alspach2019soft}
Alex Alspach, Kunimatsu Hashimoto, Naveen Kuppuswamy, and Russ Tedrake.
\newblock Soft-bubble: A highly compliant dense geometry tactile sensor for
  robot manipulation.
\newblock In \emph{2019 2nd IEEE International Conference on Soft Robotics
  (RoboSoft)}, pages 597--604. IEEE, 2019.

\bibitem[Arunachalam et~al.(2022{\natexlab{a}})Arunachalam, Güzey, Chintala,
  and Pinto]{holodex}
Sridhar~Pandian Arunachalam, Irmak Güzey, Soumith Chintala, and Lerrel Pinto.
\newblock Holo-dex: Teaching dexterity with immersive mixed reality,
  2022{\natexlab{a}}.
\newblock URL \url{https://arxiv.org/abs/2210.06463}.

\bibitem[Arunachalam et~al.(2022{\natexlab{b}})Arunachalam, Silwal, Evans, and
  Pinto]{arunachalam2022dexterous}
Sridhar~Pandian Arunachalam, Sneha Silwal, Ben Evans, and Lerrel Pinto.
\newblock Dexterous imitation made easy: A learning-based framework for
  efficient dexterous manipulation.
\newblock \emph{arXiv preprint arXiv:2203.13251}, 2022{\natexlab{b}}.

\bibitem[Bardes et~al.(2021)Bardes, Ponce, and LeCun]{bardes2021vicreg}
Adrien Bardes, Jean Ponce, and Yann LeCun.
\newblock Vicreg: Variance-invariance-covariance regularization for
  self-supervised learning.
\newblock \emph{arXiv preprint arXiv:2105.04906}, 2021.

\bibitem[Bhirangi et~al.(2021)Bhirangi, Hellebrekers, Majidi, and
  Gupta]{bhirangi2021reskin}
Raunaq~M. Bhirangi, Tess~Lee Hellebrekers, Carmel Majidi, and Abhinav Gupta.
\newblock Reskin: versatile, replaceable, lasting tactile skins.
\newblock \emph{CoRR}, abs/2111.00071, 2021.
\newblock URL \url{https://arxiv.org/abs/2111.00071}.

\bibitem[Brandfonbrener et~al.(2022)Brandfonbrener, Tu, Singh, Welker, Boodoo,
  Matni, and Varley]{https://doi.org/10.48550/arxiv.2210.02343}
David Brandfonbrener, Stephen Tu, Avi Singh, Stefan Welker, Chad Boodoo,
  Nikolai Matni, and Jake Varley.
\newblock Visual backtracking teleoperation: A data collection protocol for
  offline image-based reinforcement learning, 2022.
\newblock URL \url{https://arxiv.org/abs/2210.02343}.

\bibitem[Calandra et~al.(2018)Calandra, Owens, Jayaraman, Lin, Yuan, Malik,
  Adelson, and Levine]{calandra2018more}
Roberto Calandra, Andrew Owens, Dinesh Jayaraman, Justin Lin, Wenzhen Yuan,
  Jitendra Malik, Edward~H Adelson, and Sergey Levine.
\newblock More than a feeling: Learning to grasp and regrasp using vision and
  touch.
\newblock \emph{IEEE Robotics and Automation Letters}, 3\penalty0 (4):\penalty0
  3300--3307, 2018.

\bibitem[Caron et~al.(2020)Caron, Misra, Mairal, Goyal, Bojanowski, and
  Joulin]{caron2020unsupervised}
Mathilde Caron, Ishan Misra, Julien Mairal, Priya Goyal, Piotr Bojanowski, and
  Armand Joulin.
\newblock Unsupervised learning of visual features by contrasting cluster
  assignments.
\newblock \emph{Advances in neural information processing systems},
  33:\penalty0 9912--9924, 2020.

\bibitem[Chen et~al.(2020{\natexlab{a}})Chen, Sax, Lewis, Armeni, Savarese,
  Zamir, Malik, and Pinto]{chen2020robust}
Bryan Chen, Alexander Sax, Gene Lewis, Iro Armeni, Silvio Savarese, Amir Zamir,
  Jitendra Malik, and Lerrel Pinto.
\newblock Robust policies via mid-level visual representations: An experimental
  study in manipulation and navigation.
\newblock \emph{arXiv preprint arXiv:2011.06698}, 2020{\natexlab{a}}.

\bibitem[Chen et~al.(2021)Chen, Xu, and Agrawal]{chen2021system}
Tao Chen, Jie Xu, and Pulkit Agrawal.
\newblock A system for general in-hand object re-orientation.
\newblock \emph{Conference on Robot Learning}, 2021.

\bibitem[Chen et~al.(2020{\natexlab{b}})Chen, Kornblith, Norouzi, and
  Hinton]{simclr}
Ting Chen, Simon Kornblith, Mohammad Norouzi, and Geoffrey Hinton.
\newblock A simple framework for contrastive learning of visual
  representations.
\newblock \emph{arXiv preprint arXiv:2002.05709}, 2020{\natexlab{b}}.

\bibitem[Ciocarlie et~al.(2007)Ciocarlie, Goldfeder, and
  Allen]{Ciocarlie2007DexterousGV}
Matei~T. Ciocarlie, Corey Goldfeder, and Peter~K. Allen.
\newblock Dexterous grasping via eigengrasps : A low-dimensional approach to a
  high-complexity problem.
\newblock In \emph{Dexterous Grasping via Eigengrasps : A Low-dimensional
  Approach to a High-complexity Problem}, 2007.

\bibitem[Cui et~al.(2022)Cui, Wang, Muhammad, Pinto, et~al.]{cui2022play}
Zichen~Jeff Cui, Yibin Wang, Nur Muhammad, Lerrel Pinto, et~al.
\newblock From play to policy: Conditional behavior generation from uncurated
  robot data.
\newblock \emph{arXiv preprint arXiv:2210.10047}, 2022.

\bibitem[Dahiya et~al.(2019)Dahiya, Yogeswaran, Liu, Manjakkal, Burdet,
  Hayward, and Jörntell]{8858052}
Ravinder Dahiya, Nivasan Yogeswaran, Fengyuan Liu, Libu Manjakkal, Etienne
  Burdet, Vincent Hayward, and Henrik Jörntell.
\newblock Large-area soft e-skin: The challenges beyond sensor designs.
\newblock \emph{Proceedings of the IEEE}, 107\penalty0 (10):\penalty0
  2016--2033, 2019.
\newblock \doi{10.1109/JPROC.2019.2941366}.

\bibitem[Deng et~al.(2009)Deng, Dong, Socher, Li, Li, and
  Fei-Fei]{deng2009imagenet}
Jia Deng, Wei Dong, Richard Socher, Li-Jia Li, Kai Li, and Li~Fei-Fei.
\newblock Imagenet: A large-scale hierarchical image database.
\newblock In \emph{2009 IEEE conference on computer vision and pattern
  recognition}, pages 248--255. Ieee, 2009.

\bibitem[Dong et~al.(2017)Dong, Yuan, and Adelson]{dong2017improved}
Siyuan Dong, Wenzhen Yuan, and Edward~H Adelson.
\newblock Improved gelsight tactile sensor for measuring geometry and slip.
\newblock In \emph{2017 IEEE/RSJ International Conference on Intelligent Robots
  and Systems (IROS)}, pages 137--144. IEEE, 2017.

\bibitem[Ericsson et~al.(2022)Ericsson, Gouk, Loy, and
  Hospedales]{ericsson2022self}
Linus Ericsson, Henry Gouk, Chen~Change Loy, and Timothy~M Hospedales.
\newblock Self-supervised representation learning: Introduction, advances, and
  challenges.
\newblock \emph{IEEE Signal Processing Magazine}, 39\penalty0 (3):\penalty0
  42--62, 2022.

\bibitem[Fagard et~al.(2018)Fagard, Esseily, Jacquey, O'Regan, and
  Somogyi]{Fagard2018}
Jaqueline Fagard, Rana Esseily, Lisa Jacquey, Kevin O'Regan, and Eszter
  Somogyi.
\newblock Fetal origin of sensorimotor behavior.
\newblock \emph{Frontiers in Neurorobotics}, 12, May 2018.
\newblock \doi{10.3389/fnbot.2018.00023}.
\newblock URL \url{https://doi.org/10.3389/fnbot.2018.00023}.

\bibitem[Finn et~al.(2016)Finn, Tan, Duan, Darrell, Levine, and
  Abbeel]{finn2016deep}
Chelsea Finn, Xin~Yu Tan, Yan Duan, Trevor Darrell, Sergey Levine, and Pieter
  Abbeel.
\newblock Deep spatial autoencoders for visuomotor learning.
\newblock In \emph{2016 IEEE International Conference on Robotics and
  Automation (ICRA)}, pages 512--519. IEEE, 2016.

\bibitem[Florence et~al.(2022)Florence, Lynch, Zeng, Ramirez, Wahid, Downs,
  Wong, Lee, Mordatch, and Tompson]{florence2022implicit}
Pete Florence, Corey Lynch, Andy Zeng, Oscar~A Ramirez, Ayzaan Wahid, Laura
  Downs, Adrian Wong, Johnny Lee, Igor Mordatch, and Jonathan Tompson.
\newblock Implicit behavioral cloning.
\newblock In \emph{Conference on Robot Learning}, pages 158--168. PMLR, 2022.

\bibitem[Florence et~al.(2018)Florence, Manuelli, and
  Tedrake]{florence2018dense}
Peter~R Florence, Lucas Manuelli, and Russ Tedrake.
\newblock Dense object nets: Learning dense visual object descriptors by and
  for robotic manipulation.
\newblock In \emph{Conference on Robot Learning}, pages 373--385. PMLR, 2018.

\bibitem[Grill et~al.(2020)Grill, Strub, Altch{\'e}, Tallec, Richemond,
  Buchatskaya, Doersch, Avila~Pires, Guo, Gheshlaghi~Azar,
  et~al.]{grill2020bootstrap}
Jean-Bastien Grill, Florian Strub, Florent Altch{\'e}, Corentin Tallec, Pierre
  Richemond, Elena Buchatskaya, Carl Doersch, Bernardo Avila~Pires, Zhaohan
  Guo, Mohammad Gheshlaghi~Azar, et~al.
\newblock Bootstrap your own latent-a new approach to self-supervised learning.
\newblock \emph{NeurIPS}, 2020.

\bibitem[Ha and Schmidhuber(2018)]{ha2018world}
David Ha and J{\"u}rgen Schmidhuber.
\newblock World models.
\newblock \emph{arXiv preprint arXiv:1803.10122}, 2018.

\bibitem[Handa et~al.(2020)Handa, Van~Wyk, Yang, Liang, Chao, Wan, Birchfield,
  Ratliff, and Fox]{DexPilot}
Ankur Handa, Karl Van~Wyk, Wei Yang, Jacky Liang, Yu-Wei Chao, Qian Wan, Stan
  Birchfield, Nathan Ratliff, and Dieter Fox.
\newblock Dexpilot: Vision-based teleoperation of dexterous robotic hand-arm
  system.
\newblock In \emph{2020 IEEE International Conference on Robotics and
  Automation (ICRA)}, pages 9164--9170, 2020.
\newblock \doi{10.1109/ICRA40945.2020.9197124}.

\bibitem[Handa et~al.(2022)Handa, Allshire, Makoviychuk, Petrenko, Singh, Liu,
  Makoviichuk, Van~Wyk, Zhurkevich, Sundaralingam, Narang, Lafleche, Fox, and
  State]{nvidia2022dextreme}
Ankur Handa, Arthur Allshire, Viktor Makoviychuk, Aleksei Petrenko, Ritvik
  Singh, Jingzhou Liu, Denys Makoviichuk, Karl Van~Wyk, Alexander Zhurkevich,
  Balakumar Sundaralingam, Yashraj Narang, Jean-Francois Lafleche, Dieter Fox,
  and Gavriel State.
\newblock Dextreme: Transfer of agile in-hand manipulation from simulation to
  reality.
\newblock \emph{arXiv}, 2022.

\bibitem[He et~al.(2016)He, Zhang, Ren, and Sun]{he2016deep}
Kaiming He, Xiangyu Zhang, Shaoqing Ren, and Jian Sun.
\newblock Deep residual learning for image recognition.
\newblock In \emph{Proceedings of the IEEE conference on computer vision and
  pattern recognition}, pages 770--778, 2016.

\bibitem[Hoch et~al.(2018)Hoch, O{\textquotesingle}Grady, and Adolph]{Hoch2018}
Justine~E. Hoch, Sinclaire~M. O{\textquotesingle}Grady, and Karen~E. Adolph.
\newblock It{\textquotesingle}s the journey, not the destination: Locomotor
  exploration in infants.
\newblock \emph{Developmental Science}, 22\penalty0 (2), October 2018.
\newblock \doi{10.1111/desc.12740}.
\newblock URL \url{https://doi.org/10.1111/desc.12740}.

\bibitem[Huang et~al.(2021)Huang, Mordatch, Abbeel, and
  Pathak]{huang2021generalization}
Wenlong Huang, Igor Mordatch, Pieter Abbeel, and Deepak Pathak.
\newblock Generalization in dexterous manipulation via geometry-aware
  multi-task learning.
\newblock \emph{arXiv preprint arXiv:2111.03062}, 2021.

\bibitem[Ishihara et~al.(2006)Ishihara, Namiki, Ishikawa, and
  Shimojo]{ishihara2006dynamic}
Tatsuya Ishihara, Akio Namiki, Masatoshi Ishikawa, and Makoto Shimojo.
\newblock Dynamic pen spinning using a high-speed multifingered hand with
  high-speed tactile sensor.
\newblock In \emph{2006 6th IEEE-RAS International Conference on Humanoid
  Robots}, pages 258--263. IEEE, 2006.

\bibitem[Johansson et~al.(1992)Johansson, H{\"a}ger, and
  B{\"a}ckstr{\"o}m]{johansson1992somatosensory}
Roland~S Johansson, Charlotte H{\"a}ger, and Lars B{\"a}ckstr{\"o}m.
\newblock Somatosensory control of precision grip during unpredictable pulling
  loads: Iii. impairments during digital anesthesia.
\newblock \emph{Experimental brain research}, 89:\penalty0 204--213, 1992.

\bibitem[Kelestemur et~al.(2022)Kelestemur, Platt, and
  Padir]{kelestemur2022tactile}
Tarik Kelestemur, Robert Platt, and Taskin Padir.
\newblock Tactile pose estimation and policy learning for unknown object
  manipulation.
\newblock \emph{arXiv preprint arXiv:2203.10685}, 2022.

\bibitem[Kerr et~al.(2022)Kerr, Huang, Wilcox, Hoque, Ichnowski, Calandra, and
  Goldberg]{https://doi.org/10.48550/arxiv.2209.13042}
Justin Kerr, Huang Huang, Albert Wilcox, Ryan Hoque, Jeffrey Ichnowski, Roberto
  Calandra, and Ken Goldberg.
\newblock Learning self-supervised representations from vision and touch for
  active sliding perception of deformable surfaces, 2022.
\newblock URL \url{https://arxiv.org/abs/2209.13042}.

\bibitem[Krizhevsky et~al.(2012)Krizhevsky, Sutskever, and
  Hinton]{krizhevsky2012imagenet}
Alex Krizhevsky, Ilya Sutskever, and Geoffrey~E Hinton.
\newblock Imagenet classification with deep convolutional neural networks.
\newblock In \emph{Advances in neural information processing systems}, pages
  1097--1105, 2012.

\bibitem[Kumar et~al.(2014)Kumar, Tassa, Erez, and Todorov]{kumar2014real}
Vikash Kumar, Yuval Tassa, Tom Erez, and Emanuel Todorov.
\newblock Real-time behaviour synthesis for dynamic hand-manipulation.
\newblock In \emph{2014 IEEE International Conference on Robotics and
  Automation (ICRA)}, pages 6808--6815. IEEE, 2014.

\bibitem[Kumar et~al.(2016)Kumar, Todorov, and Levine]{kumar2016dext}
Vikash Kumar, Emanuel Todorov, and Sergey Levine.
\newblock Optimal control with learned local models: Application to dexterous
  manipulation.
\newblock In \emph{2016 IEEE International Conference on Robotics and
  Automation (ICRA)}, pages 378--383, 2016.
\newblock \doi{10.1109/ICRA.2016.7487156}.

\bibitem[Laskin et~al.(2020)Laskin, Srinivas, and Abbeel]{laskin2020curl}
Michael Laskin, Aravind Srinivas, and Pieter Abbeel.
\newblock Curl: Contrastive unsupervised representations for reinforcement
  learning.
\newblock In \emph{International Conference on Machine Learning}, pages
  5639--5650. PMLR, 2020.

\bibitem[Lee et~al.(2020)Lee, Park, Serhat, Sun, and
  Kuchenbecker]{lee2020calibrating}
Hyosang Lee, Hyunkyu Park, Gokhan Serhat, Huanbo Sun, and Katherine~J
  Kuchenbecker.
\newblock Calibrating a soft ert-based tactile sensor with a multiphysics model
  and sim-to-real transfer learning.
\newblock In \emph{2020 IEEE International Conference on Robotics and
  Automation (ICRA)}, pages 1632--1638. IEEE, 2020.

\bibitem[Lee et~al.(2019)Lee, Zhu, Zachares, Tan, Srinivasan, Savarese,
  Fei-Fei, Garg, and Bohg]{https://doi.org/10.48550/arxiv.1907.13098}
Michelle~A. Lee, Yuke Zhu, Peter Zachares, Matthew Tan, Krishnan Srinivasan,
  Silvio Savarese, Li~Fei-Fei, Animesh Garg, and Jeannette Bohg.
\newblock Making sense of vision and touch: Learning multimodal representations
  for contact-rich tasks, 2019.
\newblock URL \url{https://arxiv.org/abs/1907.13098}.

\bibitem[Lowrey et~al.(2018{\natexlab{a}})Lowrey, Kolev, Dao, Rajeswaran, and
  Todorov]{lowrey2018reinforcement}
Kendall Lowrey, Svetoslav Kolev, Jeremy Dao, Aravind Rajeswaran, and Emanuel
  Todorov.
\newblock Reinforcement learning for non-prehensile manipulation: Transfer from
  simulation to physical system.
\newblock In \emph{2018 IEEE International Conference on Simulation, Modeling,
  and Programming for Autonomous Robots (SIMPAR)}, pages 35--42. IEEE,
  2018{\natexlab{a}}.

\bibitem[Lowrey et~al.(2018{\natexlab{b}})Lowrey, Rajeswaran, Kakade, Todorov,
  and Mordatch]{lowrey2018plan}
Kendall Lowrey, Aravind Rajeswaran, Sham Kakade, Emanuel Todorov, and Igor
  Mordatch.
\newblock Plan online, learn offline: Efficient learning and exploration via
  model-based control.
\newblock \emph{arXiv preprint arXiv:1811.01848}, 2018{\natexlab{b}}.

\bibitem[Lynch et~al.(2019)Lynch, Khansari, Xiao, Kumar, Tompson, Levine, and
  Sermanet]{lynch2019learning}
Corey Lynch, Mohi Khansari, Ted Xiao, Vikash Kumar, Jonathan Tompson, Sergey
  Levine, and Pierre Sermanet.
\newblock Learning latent plans from play, 2019.

\bibitem[Mandlekar et~al.(2021)Mandlekar, Xu, Wong, Nasiriany, Wang, Kulkarni,
  Fei-Fei, Savarese, Zhu, and Mart{\'\i}n-Mart{\'\i}n]{mandlekar2021matters}
Ajay Mandlekar, Danfei Xu, Josiah Wong, Soroush Nasiriany, Chen Wang, Rohun
  Kulkarni, Li~Fei-Fei, Silvio Savarese, Yuke Zhu, and Roberto
  Mart{\'\i}n-Mart{\'\i}n.
\newblock What matters in learning from offline human demonstrations for robot
  manipulation.
\newblock \emph{arXiv preprint arXiv:2108.03298}, 2021.

\bibitem[Mordatch et~al.(2012)Mordatch, Popovi{\'c}, and
  Todorov]{mordatch2012contact}
Igor Mordatch, Zoran Popovi{\'c}, and Emanuel Todorov.
\newblock Contact-invariant optimization for hand manipulation.
\newblock In \emph{Proceedings of the ACM SIGGRAPH/Eurographics symposium on
  computer animation}, pages 137--144, 2012.

\bibitem[Murali et~al.(2020)Murali, Li, Gandhi, and
  Gupta]{10.1007/978-3-030-33950-0_33}
Adithyavairavan Murali, Yin Li, Dhiraj Gandhi, and Abhinav Gupta.
\newblock Learning to grasp without seeing.
\newblock In Jing Xiao, Torsten Kr{\"o}ger, and Oussama Khatib, editors,
  \emph{Proceedings of the 2018 International Symposium on Experimental
  Robotics}, pages 375--386, Cham, 2020. Springer International Publishing.
\newblock ISBN 978-3-030-33950-0.

\bibitem[Nagabandi et~al.(2019)Nagabandi, Konoglie, Levine, and
  Kumar]{Nagabandi2019}
Anusha Nagabandi, Kurt Konoglie, Sergey Levine, and Vikash Kumar.
\newblock Deep dynamics models for learning dexterous manipulation.
\newblock \emph{arXiv}, 2019.

\bibitem[Nair et~al.(2022)Nair, Rajeswaran, Kumar, Finn, and
  Gupta]{nair2022r3m}
Suraj Nair, Aravind Rajeswaran, Vikash Kumar, Chelsea Finn, and Abhinav Gupta.
\newblock R3m: A universal visual representation for robot manipulation.
\newblock \emph{arXiv preprint arXiv:2203.12601}, 2022.

\bibitem[Niizumi et~al.(2023)Niizumi, Takeuchi, Ohishi, Harada, and
  Kashino]{niizumi2023byol-a}
Daisuke Niizumi, Daiki Takeuchi, Yasunori Ohishi, Noboru Harada, and Kunio
  Kashino.
\newblock {BYOL for Audio}: Exploring pre-trained general-purpose audio
  representations.
\newblock \emph{IEEE/ACM Transactions on Audio, Speech, and Language
  Processing}, 31:\penalty0 137–151, 2023.
\newblock ISSN 2329-9304.
\newblock \doi{10.1109/TASLP.2022.3221007}.
\newblock URL \url{http://dx.doi.org/10.1109/TASLP.2022.3221007}.

\bibitem[Odhner et~al.(2014)Odhner, Jentoft, Claffee, Corson, Tenzer, Ma,
  Buehler, Kohout, Howe, and Dollar]{odhner2014compliant}
Lael~U Odhner, Leif~P Jentoft, Mark~R Claffee, Nicholas Corson, Yaroslav
  Tenzer, Raymond~R Ma, Martin Buehler, Robert Kohout, Robert~D Howe, and
  Aaron~M Dollar.
\newblock A compliant, underactuated hand for robust manipulation.
\newblock \emph{The International Journal of Robotics Research}, 33\penalty0
  (5):\penalty0 736--752, 2014.

\bibitem[Okamura et~al.(2000)Okamura, Smaby, and Cutkosky]{okamura2000overview}
Allison~M Okamura, Niels Smaby, and Mark~R Cutkosky.
\newblock An overview of dexterous manipulation.
\newblock In \emph{Proceedings 2000 ICRA. Millennium Conference. IEEE
  International Conference on Robotics and Automation. Symposia Proceedings
  (Cat. No. 00CH37065)}, volume~1, pages 255--262. IEEE, 2000.

\bibitem[OpenAI et~al.(2019{\natexlab{a}})OpenAI, Akkaya, Andrychowicz,
  Chociej, Litwin, McGrew, Petron, Paino, Plappert, Powell, Ribas, Schneider,
  Tezak, Tworek, Welinder, Weng, Yuan, Zaremba, and Zhang]{Openai2019}
OpenAI, Ilge Akkaya, Marcin Andrychowicz, Maciek Chociej, Mateusz Litwin, Bob
  McGrew, Arthur Petron, Alex Paino, Matthias Plappert, Glenn Powell, Raphael
  Ribas, Jonas Schneider, Nikolas Tezak, Jerry Tworek, Peter Welinder, Lilian
  Weng, Qiming Yuan, Wojciech Zaremba, and Lei Zhang.
\newblock Solving rubik's cube with a robot hand.
\newblock \emph{arXiv}, 2019{\natexlab{a}}.

\bibitem[OpenAI et~al.(2019{\natexlab{b}})OpenAI, Andrychowicz, Baker, Chociej,
  Jozefowicz, McGrew, Pachocki, Petron, Plappert, Powell, Ray, Schneider,
  Sidor, Tobin, Welinder, Weng, and Zaremba]{openai2019learning}
OpenAI, Marcin Andrychowicz, Bowen Baker, Maciek Chociej, Rafal Jozefowicz, Bob
  McGrew, Jakub Pachocki, Arthur Petron, Matthias Plappert, Glenn Powell, Alex
  Ray, Jonas Schneider, Szymon Sidor, Josh Tobin, Peter Welinder, Lilian Weng,
  and Wojciech Zaremba.
\newblock Learning dexterous in-hand manipulation, 2019{\natexlab{b}}.

\bibitem[Pari et~al.(2021)Pari, Shafiullah, Arunachalam, and
  Pinto]{pari2021surprising}
Jyothish Pari, Nur~Muhammad Shafiullah, Sridhar~Pandian Arunachalam, and Lerrel
  Pinto.
\newblock The surprising effectiveness of representation learning for visual
  imitation, 2021.

\bibitem[Patel et~al.(2021)Patel, Ouyang, Romero, and Adelson]{patel2021digger}
Radhen Patel, Rui Ouyang, Branden Romero, and Edward Adelson.
\newblock Digger finger: Gelsight tactile sensor for object identification
  inside granular media.
\newblock In \emph{Experimental Robotics: The 17th International Symposium},
  pages 105--115. Springer, 2021.

\bibitem[Pinto et~al.(2016)Pinto, Gandhi, Han, Park, and Gupta]{Pinto2016}
Lerrel Pinto, Dhiraj Gandhi, Yuanfeng Han, Yong-Lae Park, and Abhinav Gupta.
\newblock The curious robot: Learning visual representations via physical
  interactions.
\newblock In \emph{ECCV}, 2016.

\bibitem[Pomerleau(1989{\natexlab{a}})]{Pomerleau1989}
Dean~A Pomerleau.
\newblock Alvinn: An autonomous land vehicle in a neural network.
\newblock In \emph{NIPS}, 1989{\natexlab{a}}.

\bibitem[Pomerleau(1989{\natexlab{b}})]{pomerleau1989alvinn}
Dean~A Pomerleau.
\newblock Alvinn: An autonomous land vehicle in a neural network.
\newblock In \emph{NeurIPS}, pages 305--313, 1989{\natexlab{b}}.

\bibitem[Radosavovic et~al.(2022)Radosavovic, Xiao, James, Abbeel, Malik, and
  Darrell]{MVP}
Ilija Radosavovic, Tete Xiao, Stephen James, Pieter Abbeel, Jitendra Malik, and
  Trevor Darrell.
\newblock Real-world robot learning with masked visual pre-training, 2022.
\newblock URL \url{https://arxiv.org/abs/2210.03109}.

\bibitem[Rajeswaran et~al.(2018)Rajeswaran, Kumar, Gupta, Vezzani, Schulman,
  Todorov, and Levine]{Rajeswaran2018}
Aravind Rajeswaran, Vikash Kumar, Abhishek Gupta, Giulia Vezzani, John
  Schulman, Emanuel Todorov, and Sergey Levine.
\newblock Learning complex dexterous manipulation with deep reinforcement
  learning and demonstrations.
\newblock In \emph{RSS}, 2018.

\bibitem[Rochat(1989)]{rochat1989object}
Philippe Rochat.
\newblock Object manipulation and exploration in 2-to 5-month-old infants.
\newblock \emph{Developmental Psychology}, 25\penalty0 (6):\penalty0 871, 1989.

\bibitem[Schott(2003)]{Schott2003}
J~M Schott.
\newblock The grasp and other primitive reflexes.
\newblock \emph{Journal of Neurology, Neurosurgery {\&}amp Psychiatry},
  74\penalty0 (5):\penalty0 558--560, May 2003.
\newblock \doi{10.1136/jnnp.74.5.558}.
\newblock URL \url{https://doi.org/10.1136/jnnp.74.5.558}.

\bibitem[Sermanet et~al.(2018)Sermanet, Lynch, Chebotar, Hsu, Jang, Schaal,
  Levine, and Brain]{sermanet2018time}
Pierre Sermanet, Corey Lynch, Yevgen Chebotar, Jasmine Hsu, Eric Jang, Stefan
  Schaal, Sergey Levine, and Google Brain.
\newblock Time-contrastive networks: Self-supervised learning from video.
\newblock In \emph{2018 IEEE international conference on robotics and
  automation (ICRA)}, pages 1134--1141. IEEE, 2018.

\bibitem[Shafiullah et~al.(2022)Shafiullah, Cui, Altanzaya, and
  Pinto]{shafiullah2022behavior}
Nur Muhammad~Mahi Shafiullah, Zichen~Jeff Cui, Ariuntuya Altanzaya, and Lerrel
  Pinto.
\newblock Behavior transformers: Cloning \$k\$ modes with one stone.
\newblock In \emph{Advances in Neural Information Processing Systems}, 2022.
\newblock URL \url{https://openreview.net/forum?id=agTr-vRQsa}.

\bibitem[She et~al.(2019)She, Wang, Dong, Sunil, Rodriguez, and
  Adelson]{https://doi.org/10.48550/arxiv.1910.02860}
Yu~She, Shaoxiong Wang, Siyuan Dong, Neha Sunil, Alberto Rodriguez, and Edward
  Adelson.
\newblock Cable manipulation with a tactile-reactive gripper, 2019.
\newblock URL \url{https://arxiv.org/abs/1910.02860}.

\bibitem[Shigemi(2018)]{Shigemi2018}
Satoshi Shigemi.
\newblock \emph{ASIMO and Humanoid Robot Research at Honda}, pages 1--36.
\newblock Springer Netherlands, Dordrecht, 2018.
\newblock ISBN 978-94-007-7194-9.
\newblock \doi{10.1007/978-94-007-7194-9_9-2}.
\newblock URL \url{https://doi.org/10.1007/978-94-007-7194-9_9-2}.

\bibitem[Sievers et~al.(2022)Sievers, Pitz, and Bäuml]{9812093}
Leon Sievers, Johannes Pitz, and Berthold Bäuml.
\newblock Learning purely tactile in-hand manipulation with a torque-controlled
  hand.
\newblock In \emph{2022 International Conference on Robotics and Automation
  (ICRA)}, pages 2745--2751, 2022.
\newblock \doi{10.1109/ICRA46639.2022.9812093}.

\bibitem[Song et~al.(2020)Song, Zeng, Lee, and Funkhouser]{song2020grasping}
Shuran Song, Andy Zeng, Johnny Lee, and Thomas Funkhouser.
\newblock Grasping in the wild: Learning 6dof closed-loop grasping from
  low-cost demonstrations.
\newblock \emph{RA-L}, 2020.

\bibitem[Tomo et~al.(2018)Tomo, Regoli, Schmitz, Natale, Kristanto, Somlor,
  Jamone, Metta, and Sugano]{8307485}
Tito~Pradhono Tomo, Massimo Regoli, Alexander Schmitz, Lorenzo Natale, Harris
  Kristanto, Sophon Somlor, Lorenzo Jamone, Giorgio Metta, and Shigeki Sugano.
\newblock A new silicone structure for uskin—a soft, distributed, digital
  3-axis skin sensor and its integration on the humanoid robot icub.
\newblock \emph{IEEE Robotics and Automation Letters}, 3\penalty0 (3):\penalty0
  2584--2591, 2018.
\newblock \doi{10.1109/LRA.2018.2812915}.

\bibitem[Wang et~al.(2018)Wang, Wu, Sun, Yuan, Freeman, Tenenbaum, and
  Adelson]{wang20183d}
Shaoxiong Wang, Jiajun Wu, Xingyuan Sun, Wenzhen Yuan, William~T Freeman,
  Joshua~B Tenenbaum, and Edward~H Adelson.
\newblock 3d shape perception from monocular vision, touch, and shape priors.
\newblock In \emph{2018 IEEE/RSJ International Conference on Intelligent Robots
  and Systems (IROS)}, pages 1606--1613. IEEE, 2018.

\bibitem[Wang et~al.(2021{\natexlab{a}})Wang, She, Romero, and
  Adelson]{https://doi.org/10.48550/arxiv.2106.08851}
Shaoxiong Wang, Yu~She, Branden Romero, and Edward Adelson.
\newblock Gelsight wedge: Measuring high-resolution 3d contact geometry with a
  compact robot finger, 2021{\natexlab{a}}.
\newblock URL \url{https://arxiv.org/abs/2106.08851}.

\bibitem[Wang et~al.(2021{\natexlab{b}})Wang, Huang, Fang, Sun, and
  Li]{wang2021elastic}
Yikai Wang, Wenbing Huang, Bin Fang, Fuchun Sun, and Chang Li.
\newblock Elastic tactile simulation towards tactile-visual perception.
\newblock In \emph{Proceedings of the 29th ACM International Conference on
  Multimedia}, pages 2690--2698, 2021{\natexlab{b}}.

\bibitem[Yarats et~al.(2021)Yarats, Fergus, Lazaric, and
  Pinto]{yarats2021reinforcement}
Denis Yarats, Rob Fergus, Alessandro Lazaric, and Lerrel Pinto.
\newblock Reinforcement learning with prototypical representations.
\newblock In \emph{International Conference on Machine Learning}, pages
  11920--11931. PMLR, 2021.

\bibitem[Yarats et~al.(2022)Yarats, Brandfonbrener, Liu, Laskin, Abbeel,
  Lazaric, and Pinto]{yarats2022don}
Denis Yarats, David Brandfonbrener, Hao Liu, Michael Laskin, Pieter Abbeel,
  Alessandro Lazaric, and Lerrel Pinto.
\newblock Don't change the algorithm, change the data: Exploratory data for
  offline reinforcement learning.
\newblock \emph{arXiv preprint arXiv:2201.13425}, 2022.

\bibitem[Young et~al.(2020)Young, Gandhi, Tulsiani, Gupta, Abbeel, and
  Pinto]{young2020visual}
Sarah Young, Dhiraj Gandhi, Shubham Tulsiani, Abhinav Gupta, Pieter Abbeel, and
  Lerrel Pinto.
\newblock Visual imitation made easy, 2020.

\bibitem[Young et~al.(2021)Young, Pari, Abbeel, and Pinto]{young2021playful}
Sarah Young, Jyothish Pari, Pieter Abbeel, and Lerrel Pinto.
\newblock Playful interactions for representation learning.
\newblock \emph{arXiv preprint arXiv:2107.09046}, 2021.

\bibitem[Zambelli et~al.(2021)Zambelli, Aytar, Visin, Zhou, and
  Hadsell]{zambelli2021learning}
Martina Zambelli, Yusuf Aytar, Francesco Visin, Yuxiang Zhou, and Raia Hadsell.
\newblock Learning rich touch representations through cross-modal
  self-supervision.
\newblock In \emph{Conference on Robot Learning}, pages 1415--1425. PMLR, 2021.

\bibitem[Zhan et~al.(2020)Zhan, Zhao, Pinto, Abbeel, and
  Laskin]{zhan2020framework}
Albert Zhan, Ruihan Zhao, Lerrel Pinto, Pieter Abbeel, and Michael Laskin.
\newblock A framework for efficient robotic manipulation.
\newblock In \emph{Deep RL Workshop NeurIPS 2021}, 2020.

\bibitem[Zhou et~al.(2022)Zhou, Dean, Srirama, Rajeswaran, Pari, Hatch, Jain,
  Yu, Abbeel, Pinto, Finn, and Gupta]{zhou2022train}
Gaoyue Zhou, Victoria Dean, Mohan~Kumar Srirama, Aravind Rajeswaran, Jyothish
  Pari, Kyle~Beltran Hatch, Aryan Jain, Tianhe Yu, Pieter Abbeel, Lerrel Pinto,
  Chelsea Finn, and Abhinav Gupta.
\newblock Train offline, test online: A real robot learning benchmark.
\newblock In \emph{Deep Reinforcement Learning Workshop NeurIPS 2022}, 2022.
\newblock URL \url{https://openreview.net/forum?id=eqrVnNgkYWZ}.

\bibitem[Zhu et~al.(2019)Zhu, Gupta, Rajeswaran, Levine, and Kumar]{Zhu2019}
Henry Zhu, Abhishek Gupta, Aravind Rajeswaran, Sergey Levine, and Vikash Kumar.
\newblock Dexterous manipulation with deep reinforcement learning: Efficient,
  general, and low-cost.
\newblock In \emph{ICRA}, 2019.

\bibitem[Zhu et~al.(2022)Zhu, Joshi, Stone, and Zhu]{zhu2022viola}
Yifeng Zhu, Abhishek Joshi, Peter Stone, and Yuke Zhu.
\newblock Viola: Imitation learning for vision-based manipulation with object
  proposal priors.
\newblock \emph{arXiv preprint arXiv:2210.11339}, 2022.

\end{thebibliography}

\clearpage
\newpage
\appendix
\label{sec:appendix}

\subsection{Experiment Details}
\label{sec:exp_details}
For simplicity, we secure the joystick, bottle, and book to the table. This mimics having another manipulator keep the object in place while the hand manipulates the object. The bowls and cups are not secured, making the problem of unstacking much more difficult.

To ensure fair evaluation, we start each method with the object in the same  configuration for each index trial. This corresponds to 10 different starting positions, each of which is used at the start of each baseline run.

\subsection{Model Details}
\label{sec:model_details}

Here we provide additional details about our method and baselines for easier reproduction.

For all image-based models, we normalize the inputs based on the mean and standard deviation of the data seen during training. For the tactile-based models, we normalize the inputs to be within the range $[0,1]$.

\subsubsection{BYOL Details}
The complete list of BYOL hyperparameters has been provided in Table~\ref{tab:hyperparams}. We take the model with the lowest training loss out of all the epochs.

\begin{table}[h]
    \begin{center}
    \setlength{\tabcolsep}{6pt}
    \renewcommand{\arraystretch}{1.5}
    \begin{tabular}{ c c } 
        \hline
        Parameter & Value \\
        \hline
                Optimizer          & Adam\\
                Learning rate      & $1e^{-3}$\\
                Weight decay   & $1e^{-5}$\\
                Max epochs &  1000\\
                Batch size (Tactile) & 1024 \\
                Batch size (Image) & 64 \\
                Aug. (Tactile) & Gaussian Blur (3x3) (1.0, 2.0) \ $p=0.5$ \\
                                    & Random Resize Crop (0.9, 1.0)  \ $p=0.5$ \\
                Aug. (Image) & Color Jitter (0.8, 0.8, 0.8, 0.2) \ $p=0.2$ \\
                & Gaussian Blur (3x3) (1.0, 2.0) \ $p=0.2$ \\
                                    & Random Grayscale \ $p=0.2$ \\
        \hline
    \end{tabular}
    \end{center}
    \caption{BYOL Hyperparameters.}
    \label{tab:hyperparams}
\end{table}

\subsubsection{Nearest Neighbors Details}
We give equal weight to visual and tactile distances for all of the tasks except bottle cap, where tactile and image features were given weights of 2 and 1, respectively. We do this because the quality of the neighbors on image data was poor and emphasizing the tactile data slightly vastly improves performance.

While executing NN imitation, we keep a buffer of recently executed neighbors that we call the reject buffer. Given a new observation, we pick the first nearest neighbor not in the reject buffer. This prevents the policy from getting stuck in loops if a chain of neighbors and actions are cyclical.
We set the reject buffer size to 10 for every task except Joystick Pulling, which is set to 3. 
The buffer, combined with the 2cm spatial subsampling are critical for the success of NN policies.

\subsubsection{BC Details}
We train BC end-to-end using standard MSE loss on the actions with the same learning rate as BYOL and a batch size of 64.
\subsubsection{NN-Torque Details}
Our hand does not have torque sensors, but is actuated by torque targets from a low-level PD position controller. We use the torque targets as a proxy for torque information since the desired torque will be higher when the finger is in contact with an object, but trying to move further inside, and lower when it is not in contact. 

\subsubsection{PCA Details}
\begin{figure}
    \centering
    \includegraphics[width=0.5\textwidth]{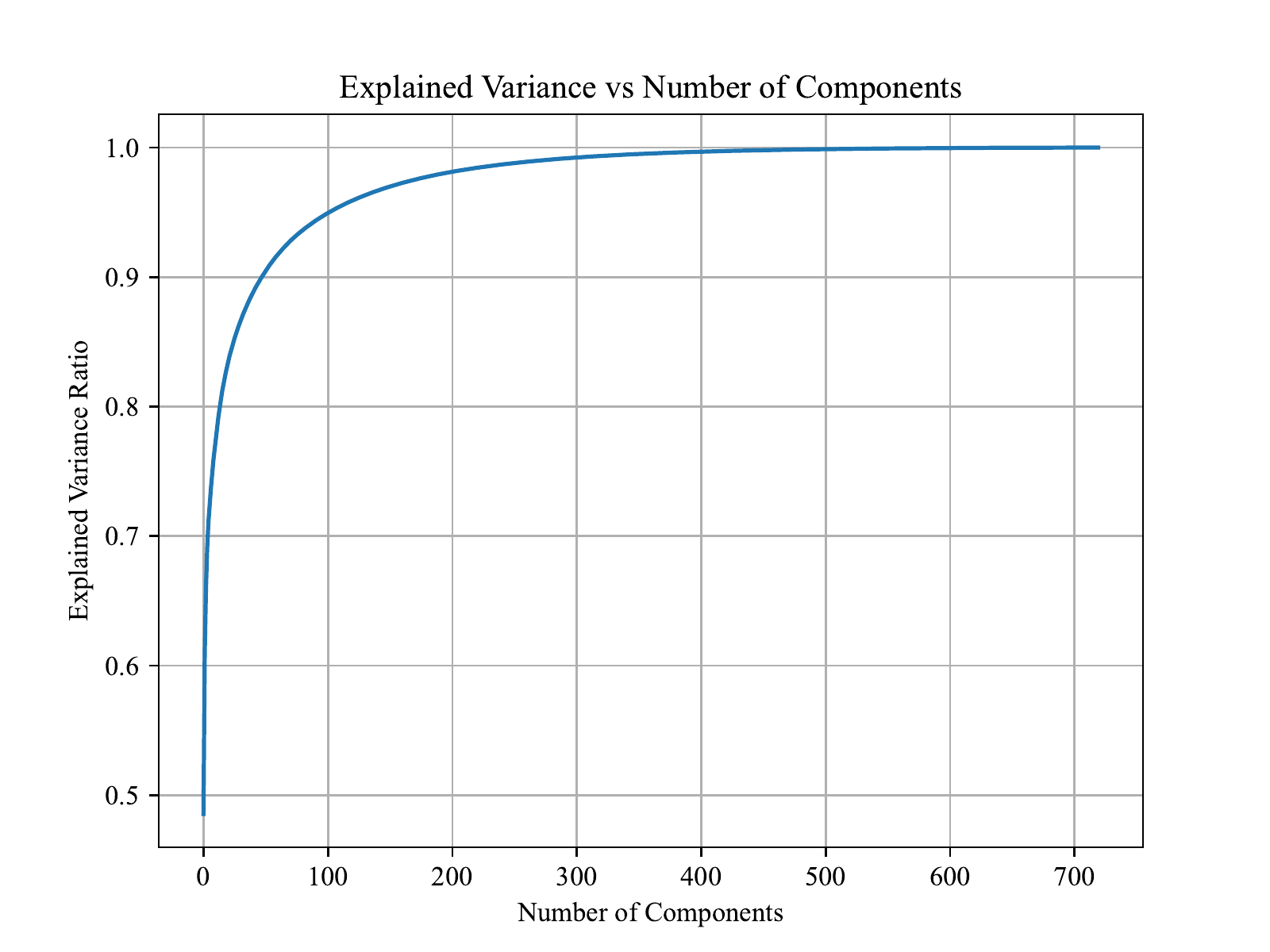}
    \caption{Explained variance ratio for PCA on the play tactile data. Most variance is captured in the first 100 components.}
    \label{fig:ap_var}
\end{figure}

\begin{figure*}
    \centering
    \begin{tabular}{@{}c@{}}
        \includegraphics[width=0.49\textwidth]{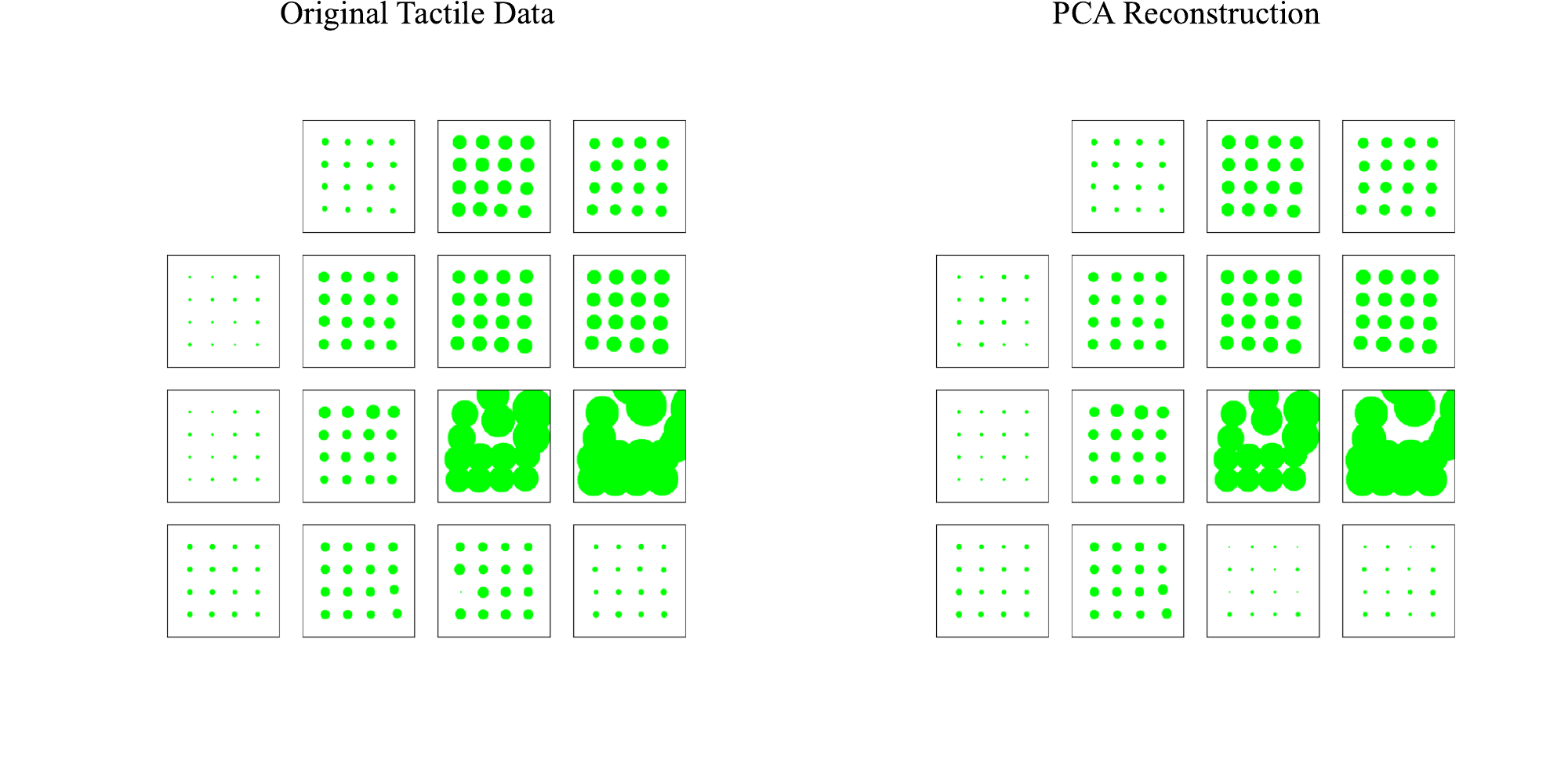}
        \label{fig:ap_pca1}
    \end{tabular}
    \begin{tabular}{@{}c@{}}
        \includegraphics[width=0.49\textwidth]{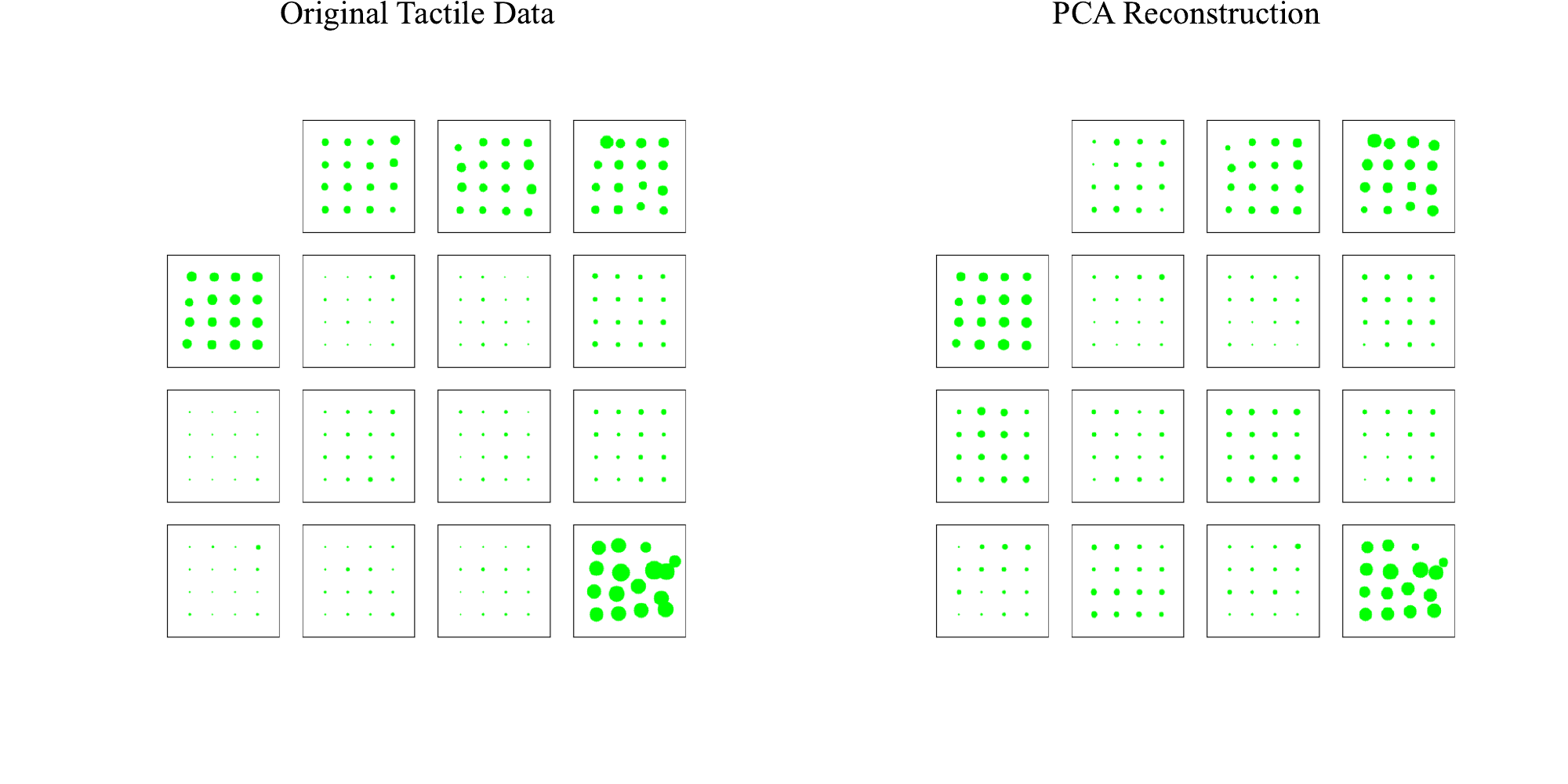}
        \label{fig:ap_pca2}
    \end{tabular}
    \caption{Tactile data and the PCA reconstruction of two using 100 components for two tactile readings. Most of the information is preserved, but we can see minor differences in magnitude and offset.}
    \label{fig:ap_pca}
\end{figure*}

We run PCA on the tactile play data and take the top 100 components for use as features. The captured variance is about 95\% and the entire explained variance ratio can be seen in Figure \ref{fig:ap_var}. By visualizing the reconstructions (Figure \ref{fig:ap_pca}), we can see that it retains a majority of the tactile information.

\subsubsection{Shuffled Pad Details}
For this experiment, we permute the position of the 15 4x4 tactile sensors using the same permutation for both pretraining and deployment. This ensures that we're inputting the same data from each sensor to each location in the tactile image, but does not leverage the spatial locations of the pads on the hand. If spatial layout had no effect, we would expect no difference in the performance between this and \method.

\subsection{Additional Rollouts}
\label{sec:ap_rollouts}
\begin{figure*}
    \centering
    \includegraphics[width=\textwidth]{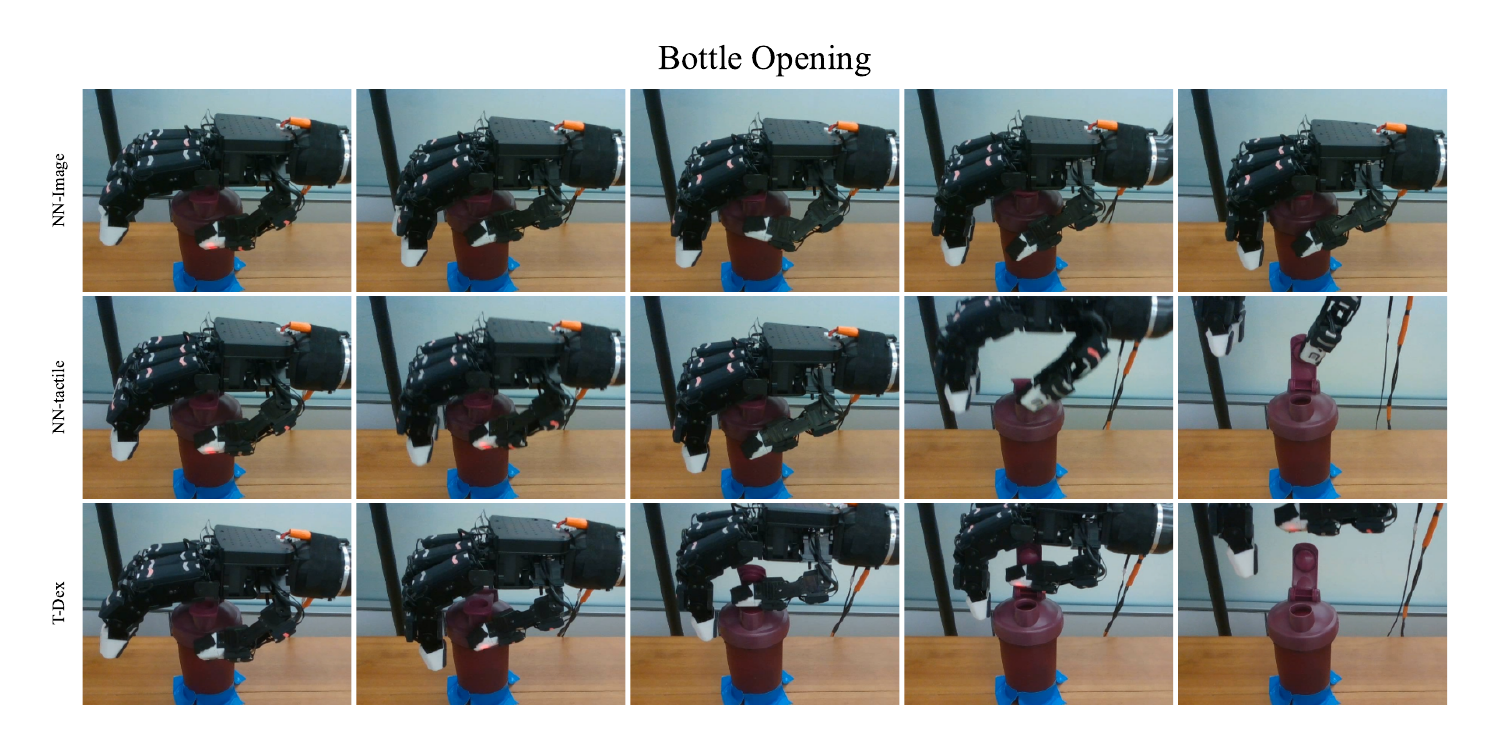}
    \caption{Additional rollouts for the Bottle Opening task.}
    \label{fig:ap_bottle}
\end{figure*}

\begin{figure*}
    \centering
    \includegraphics[width=\textwidth]{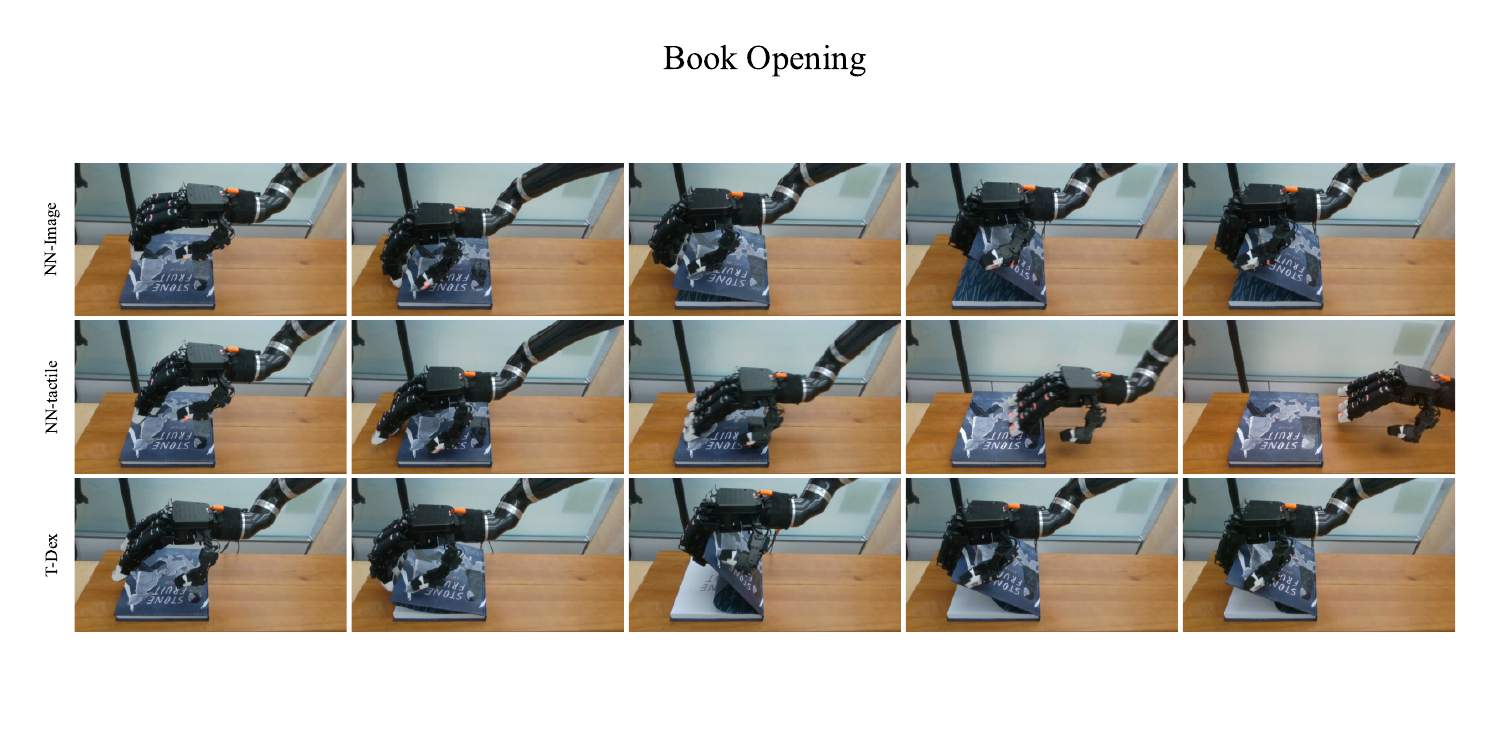}
    \caption{Additional rollouts for the Book Opening task.}
    \label{fig:ap_book}
\end{figure*}

\begin{figure*}
    \centering
    \includegraphics[width=\textwidth]{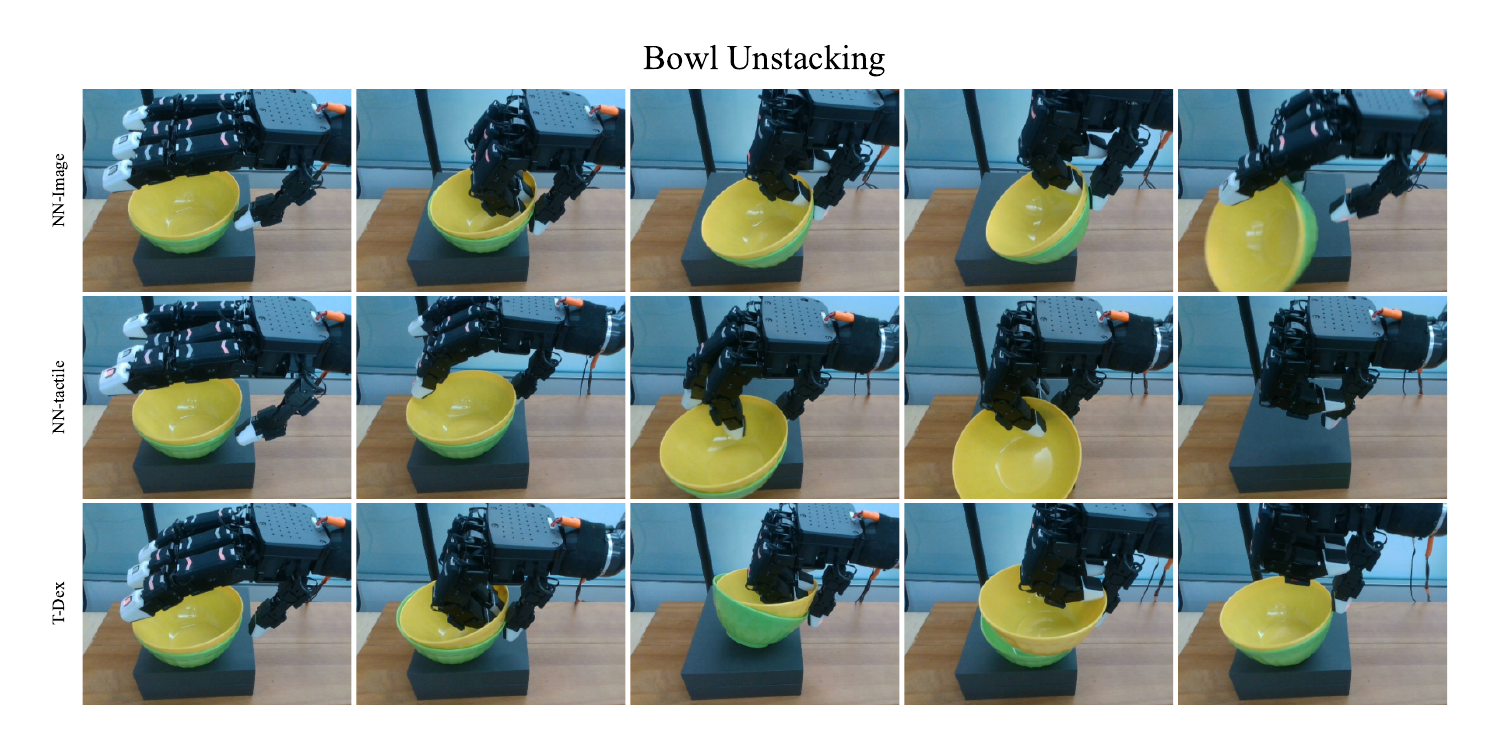}
    \caption{Additional rollouts for the Bowl Unstacking task.}
    \label{fig:ap_bowl}
\end{figure*}

\begin{figure*}
    \centering
    \includegraphics[width=\textwidth]{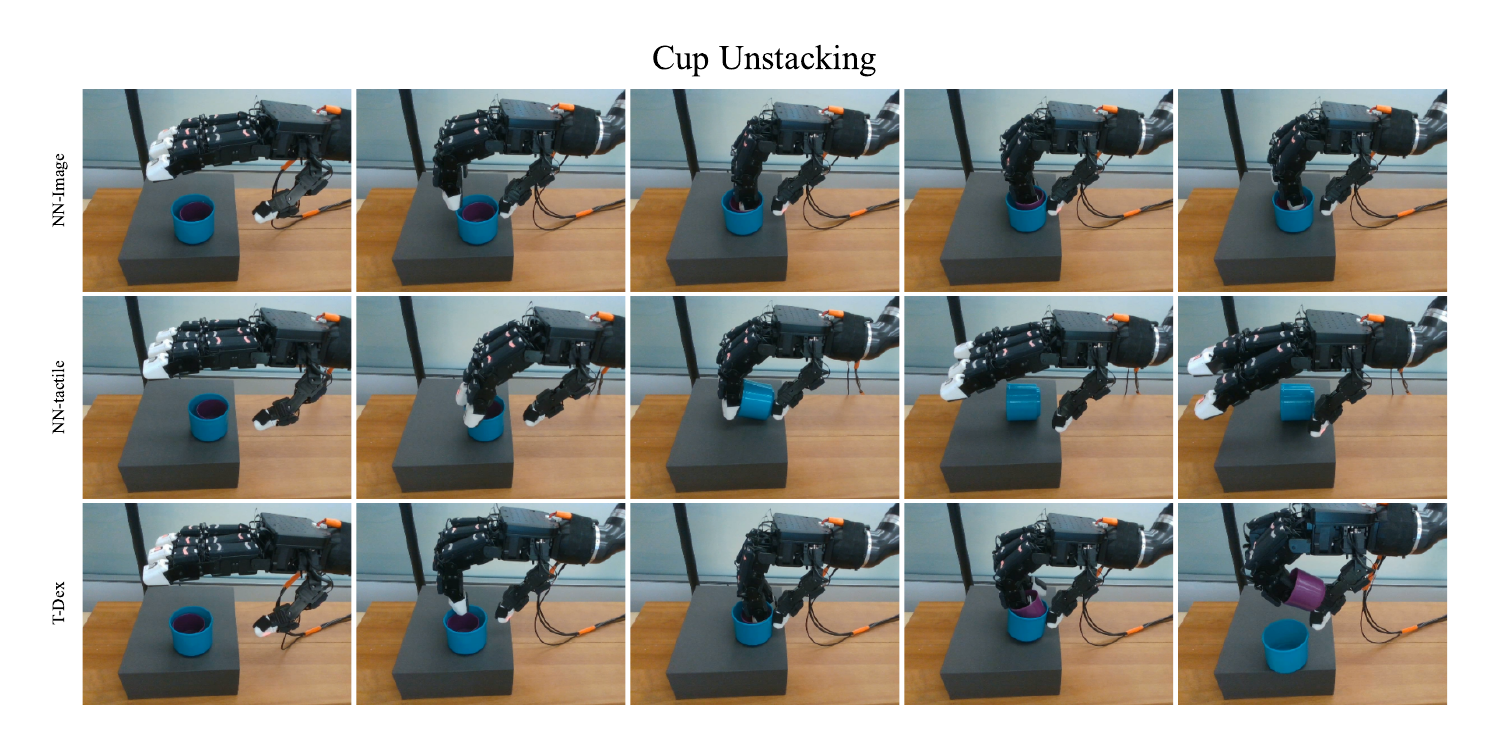}
    \caption{Additional rollouts for the Cup Unstacking task.}
    \label{fig:ap_cup}
\end{figure*}

\begin{figure*}[ht]
    \centering
    \includegraphics[width=\textwidth]{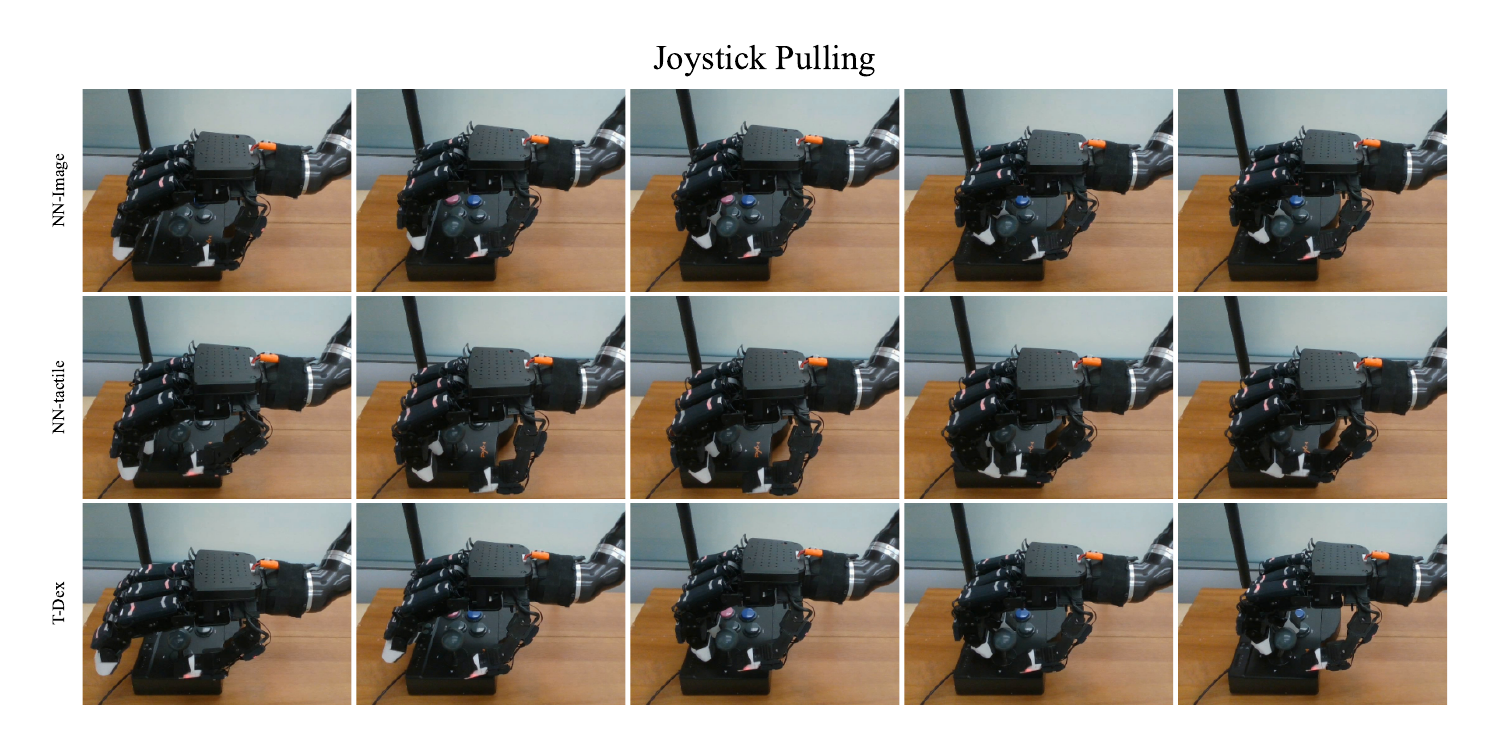}
    \caption{Additional rollouts for the Joystick Pulling task.}
    \label{fig:ap_joystick}
\end{figure*}

We visualize extra rollouts for each task in Figures \ref{fig:ap_bottle}-\ref{fig:ap_joystick}.

\subsection{Tactile Image Visualization}
\begin{figure*}
    \centering
    \includegraphics[width=0.9\textwidth]{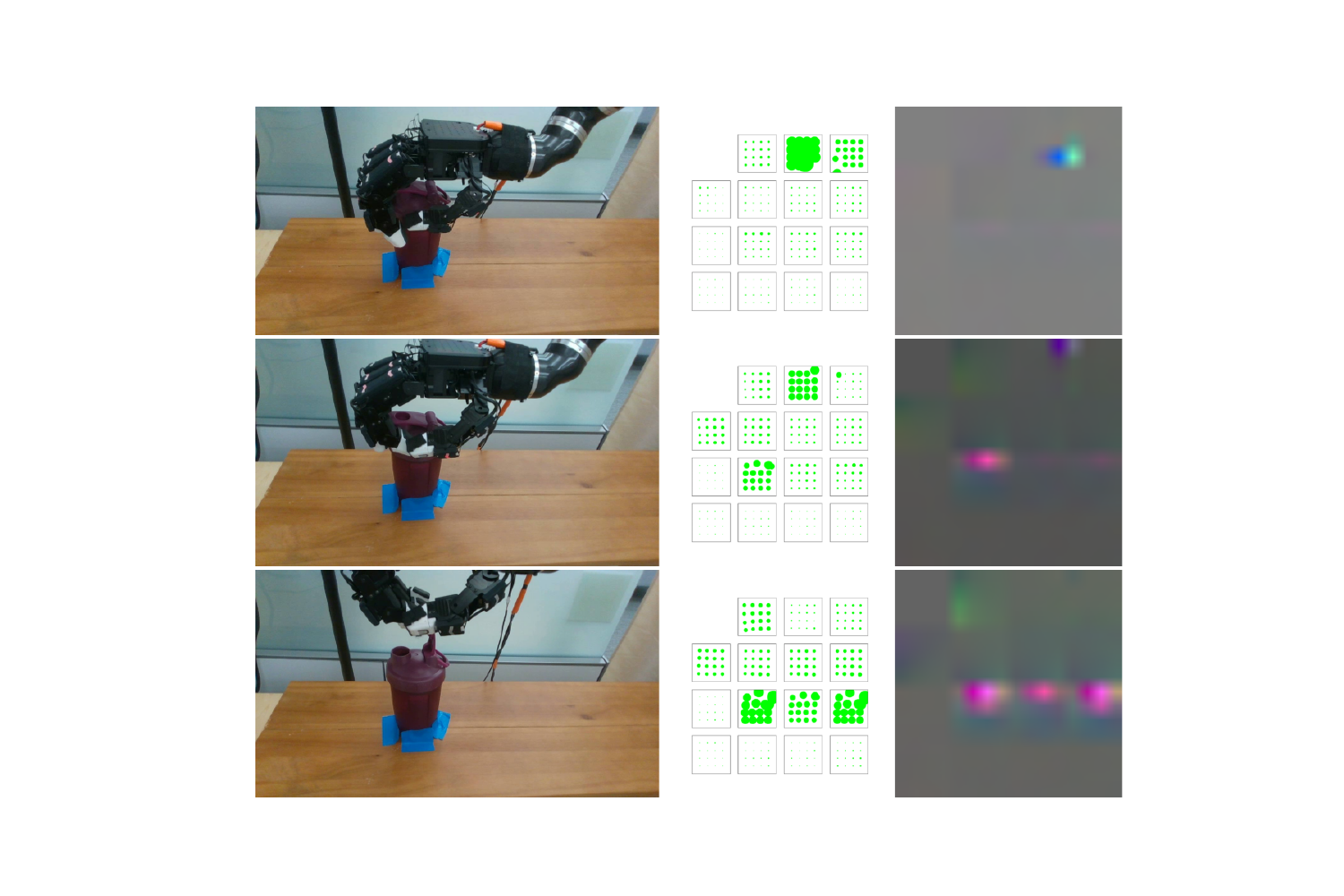}
    \caption{Tactile Image for the Bottle Opening task.}
    \label{fig:ap_tact_bottle}
\end{figure*}

\begin{figure*}
    \centering
    \includegraphics[width=0.9\textwidth]{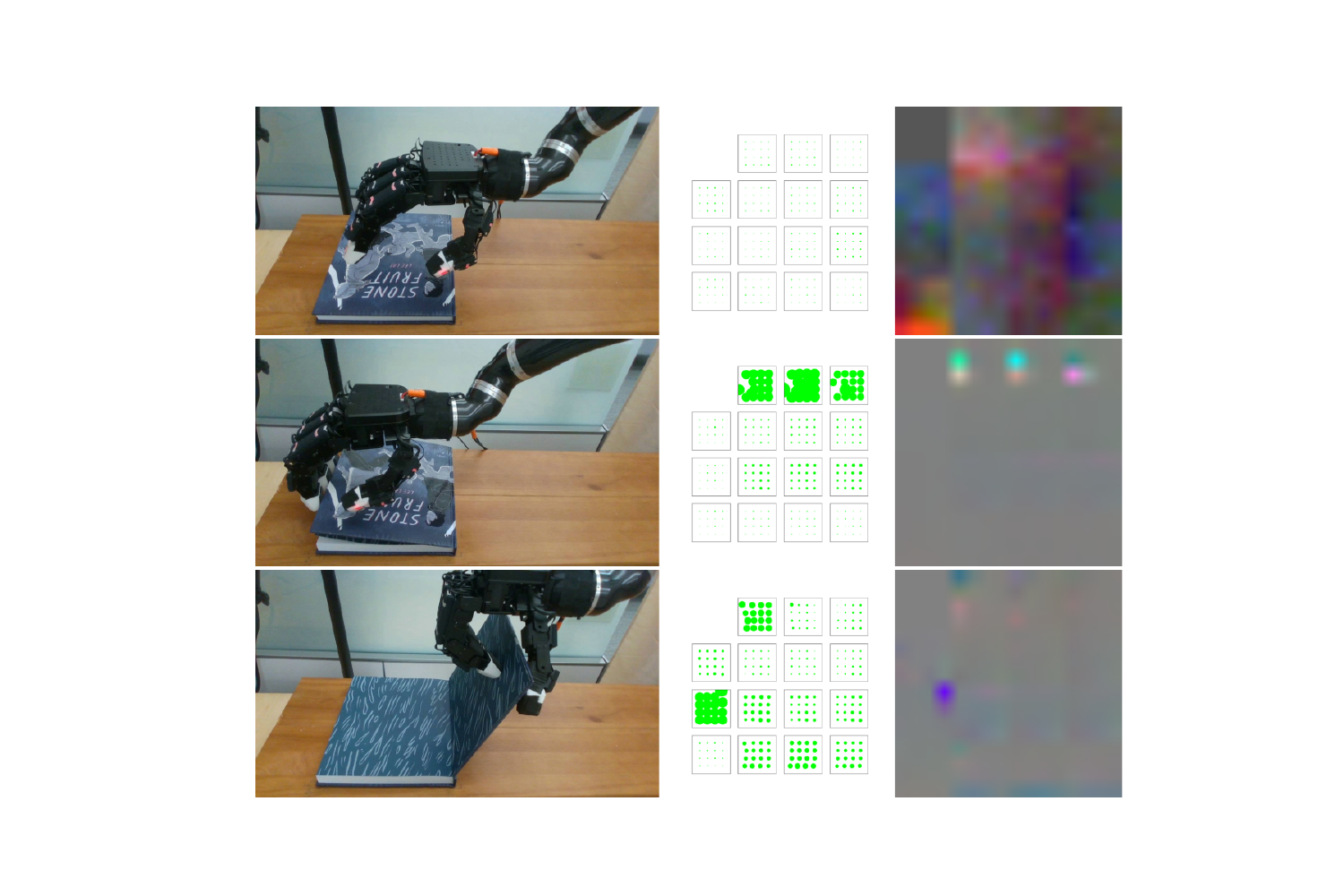}
    \caption{Tactile Image for the Book Opening task.}
    \label{fig:ap_tact_book}
\end{figure*}

\begin{figure*}
    \centering
    \includegraphics[width=0.9\textwidth]{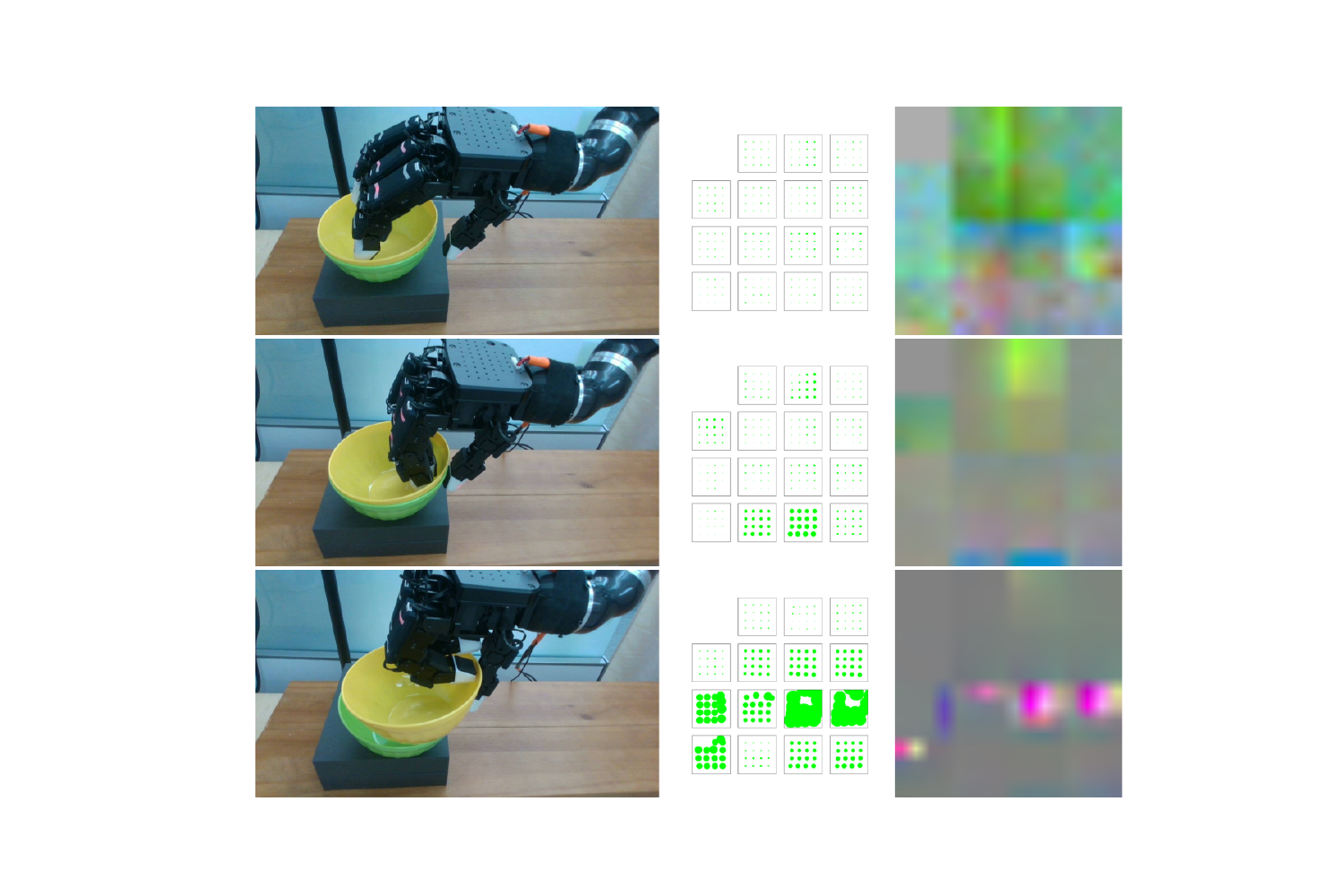}
    \caption{Tactile Image for the Bowl Unstacking task.}
    \label{fig:ap_tact_bowl}
\end{figure*}

\begin{figure*}
    \centering
    \includegraphics[width=0.9\textwidth]{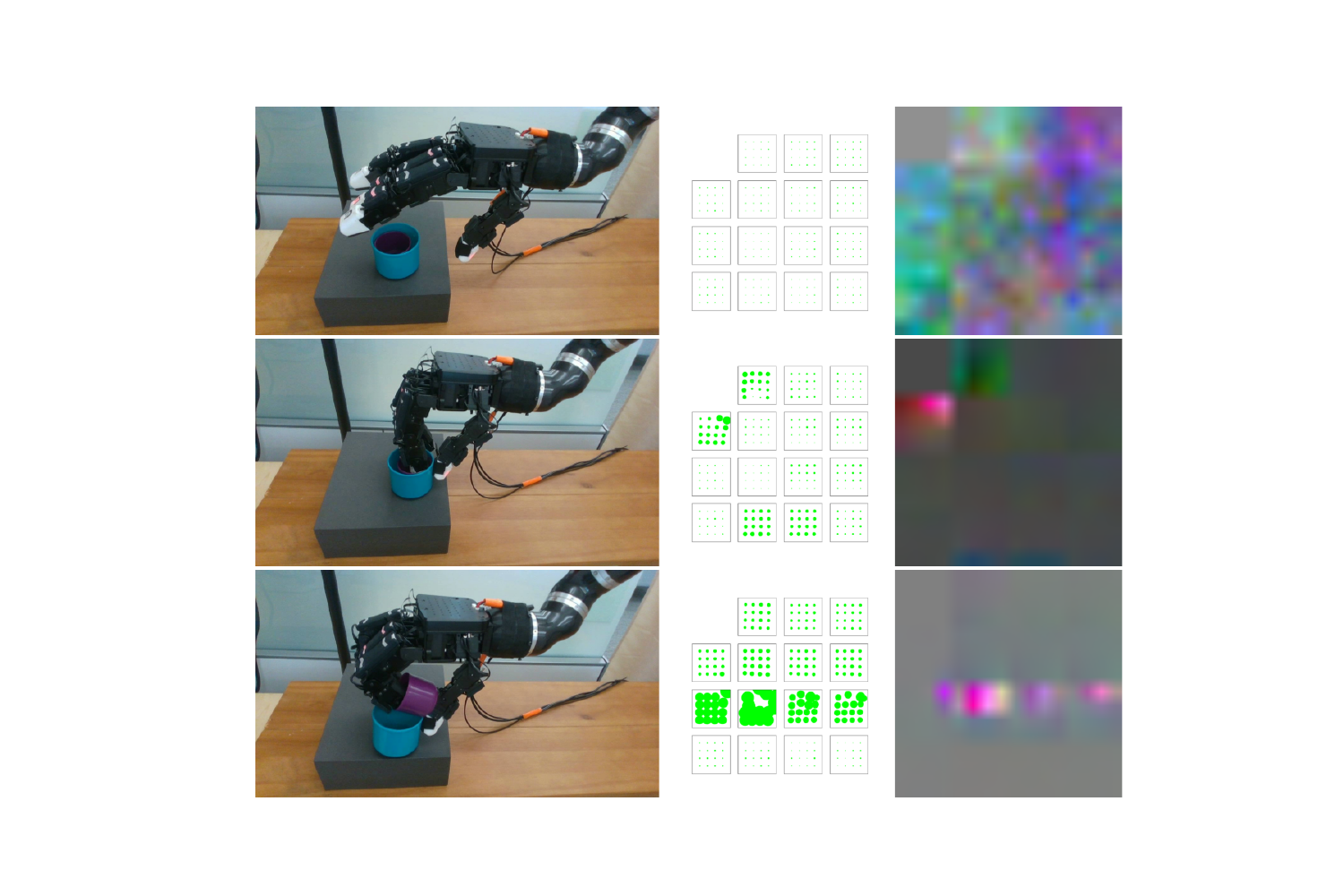}
    \caption{Tactile Image for the Cup Unstacking task.}
    \label{fig:ap_tact_cup}
\end{figure*}

\begin{figure*}
    \centering
    \includegraphics[width=0.9\textwidth]{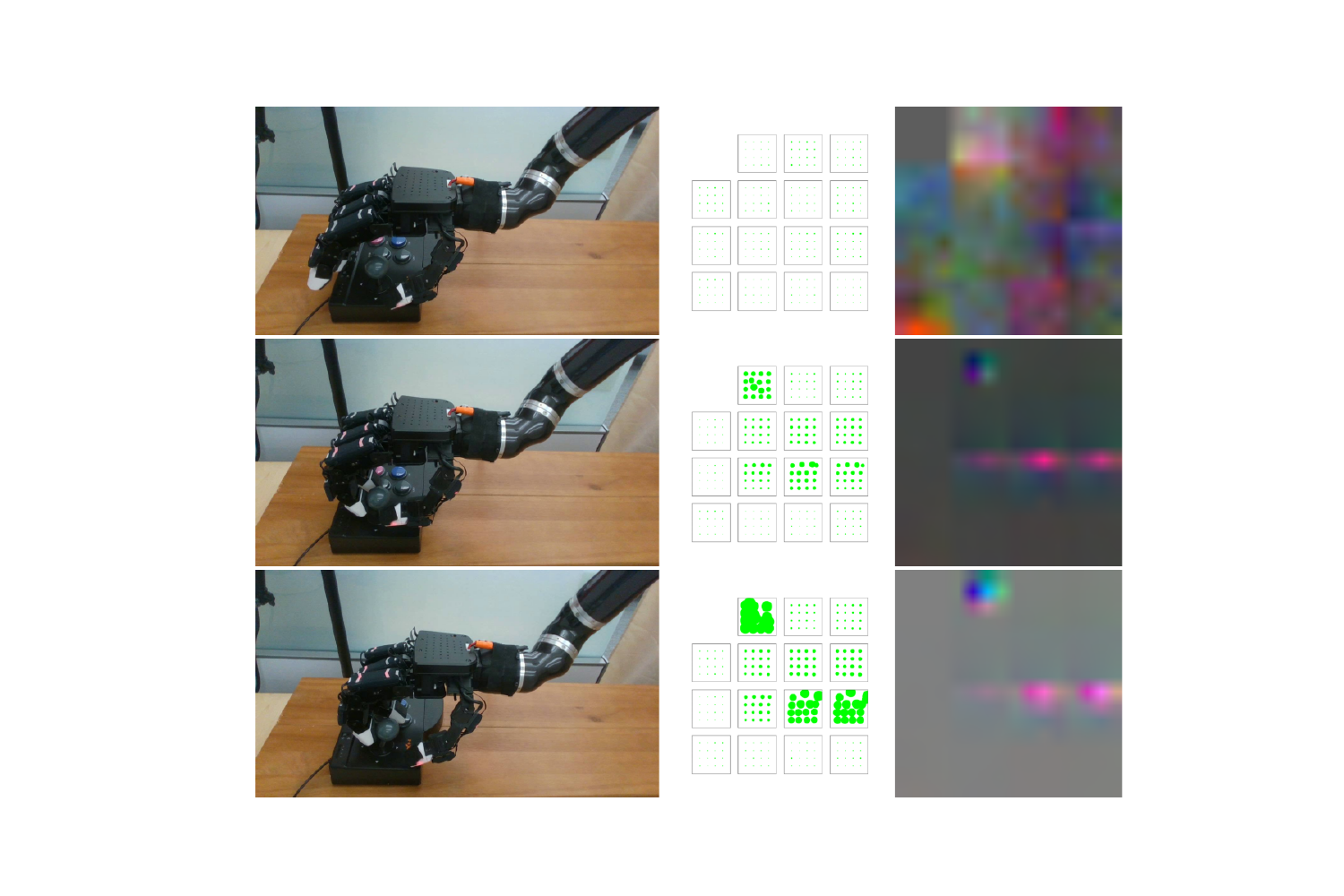}
    \caption{Tactile Image for the Joystick Pulling task.}
    \label{fig:ap_tact_joystic}
\end{figure*}

We visualize tactile images for each task in Figures \ref{fig:ap_tact_bottle}-\ref{fig:ap_tact_joystic}.

\end{document}